\documentclass{article}

\usepackage[preprint,nonatbib]{neurips_2022}

\usepackage[numbers]{natbib}
\usepackage{amssymb}
\usepackage[utf8]{inputenc} 
\usepackage[T1]{fontenc}    
\usepackage{hyperref}       
\usepackage{url}            
\usepackage{booktabs}       
\usepackage{amsfonts}       
\usepackage{nicefrac}       
\usepackage{microtype}      
\usepackage[table]{xcolor}
\usepackage{tikz}
\usepackage{xcolor}         
\usepackage{tabularx}
\usepackage{makecell}
\usepackage{natbib}

\usepackage{bm}
\usepackage{url}            
\usepackage{booktabs}       
\usepackage{amsfonts,amsmath}       
\usepackage{nicefrac}       
\usepackage{microtype}      
\usepackage{xcolor}
\usepackage{tabu}
\usepackage{multirow}       
\usepackage{graphicx}
\usepackage{graphics}
\usepackage{tabularx}
\usepackage{makecell}
\usepackage{caption}
\usepackage{wrapfig}
\usepackage{enumitem}
\usepackage{subcaption}
\usepackage{arydshln}
\usepackage{algpseudocode}
\usepackage{amsfonts}       
\usepackage{nicefrac}       
\usepackage{color, colortbl}
\definecolor{Gray}{gray}{0.95}
\definecolor{orange}{rgb}{0.9,0.5,0}

\def\Sigmoid{\textsf{Sigmoid}} 
\def\GeLU{\textsf{GeLU}} 
\def\DWConv{\textsf{DWConv}} 
\def\AvgPool{\textsf{Avg-Pool}} 

\newcommand{\eg}[0]{\emph{e.g., }}

\newcommand{\beq}{\vspace{0mm}\begin{equation}}
\newcommand{\eeq}{\vspace{0mm}\end{equation}}
\newcommand{\beqs}{\vspace{0mm}\begin{eqnarray}}
\newcommand{\eeqs}{\vspace{0mm}\end{eqnarray}}
\newcommand{\barr}{\begin{array}}
\newcommand{\earr}{\end{array}}

\newcommand{\Gmat}{{\bf G}}

\newcommand{\Mmat}{{\bf M}}

\newcommand{\Xmat}[0]{{{\bf X}}}

\newcommand{\Zmat}{{\bf Z}}

\newcommand{\gv}[0]{{\boldsymbol{g}}}

\newcommand{\xv}{\boldsymbol{x}}
\newcommand{\yv}{\boldsymbol{y}}
\newcommand{\zv}{\boldsymbol{z}}

\newcommand{\R}{\mathbb{R}}

\usepackage{color, colortbl}
\definecolor{Gray}{gray}{0.93}

\usepackage{xcolor,amsmath}
\usepackage[linesnumbered,ruled,vlined]{algorithm2e}
\DontPrintSemicolon

\newcommand{\forinline}{ \textcolor{blue!90!black} }

\SetKwComment{Comment}{\color{green!50!black}\# }{}
\renewcommand{\CommentSty}[1]{\textnormal{\ttfamily\color{green!50!black}#1}\unskip}

\newcommand{\var}{\texttt}
\newcommand{\FuncCall}[2]{\texttt{\bfseries #1(#2)}}
\SetKwProg{Function}{def}{:}{}

\SetKwProg{For}{for}{:}{}
\SetKwProg{If}{if}{:}{}

\usepackage{pifont}
\newcommand{\cmark}{\ding{51}}%
\newcommand{\xmark}{\ding{55}}%

\title{Focal Modulation Networks}

\author{%
  Jianwei Yang$^1$, Chunyuan Li$^1$, Xiyang Dai$^2$, Lu Yuan$^2$, Jianfeng Gao$^1$ \\
  \small{$^1$Microsoft Research at Redmond, $^2$Microsoft Cloud + AI} \\
  \texttt{\{jianwyan,chunyl,xidai,luyuan,jfgao\}@microsoft.com} \\
}

\begin{document}

\maketitle

\vspace{-15pt}
\begin{figure}[h]
    \centering
    \includegraphics[width=0.8\linewidth]{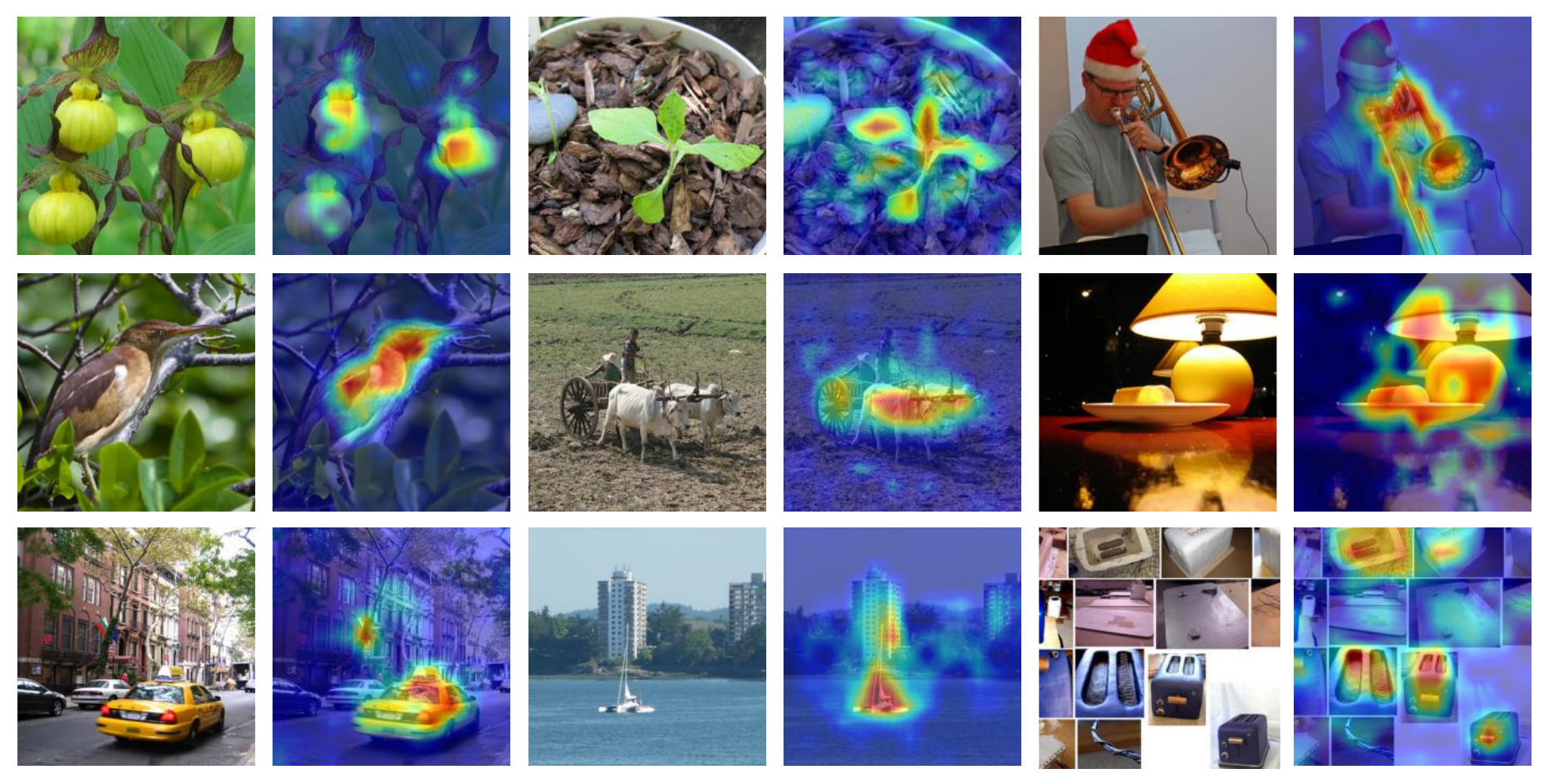}
    \caption{Given images, Focal Modulation Network (FocalNet) exhibits extraordinary interpretability -- the modulation automatically and gradually converge on the object regions that induce the recognition categories. NO visual explanation tools like CAM~\cite{zhou2016learning} or Grad-CAM~\cite{selvaraju2017grad} are used.}
    \label{fig:focal_modulation_visualization}
\end{figure}

\begin{abstract}
  We propose \textit{focal modulation networks} (\emph{FocalNets} in short), where self-attention (SA) is completely replaced by a \emph{focal modulation} module for modeling token interactions in vision. Focal modulation comprises three components: $(i)$ focal contextualization, implemented using a stack of depth-wise convolutional layers, to encode visual contexts from short to long ranges, $(ii)$ gated aggregation to selectively gather contexts into a modulator for each query token, and $(iii)$ element-wise affine transformation to inject the modulator into the query. Extensive experiments show FocalNets exhibit extraordinary interpretability (Fig.~\ref{fig:focal_modulation_visualization}) and outperform the SoTA SA counterparts (\textit{e.g.}, Swin and Focal Transformers) with similar computational cost on the tasks of image classification, object detection, and segmentation. Specifically, FocalNets with tiny and base size can achieve \textbf{82.3}\% and \textbf{83.9}\% top-1 accuracy on ImageNet-1K. After pretrained on ImageNet-22K in 224$^2$ resolution, it attains \textbf{86.5}\% and \textbf{87.3}\% top-1 accuracy when finetuned with resolution 224$^2$ and 384$^2$, respectively. For object detection with Mask R-CNN~\cite{he2017mask}, FocalNet base trained with 1$\times$ outperforms the Swin counterpart by \textbf{2.1} points and already surpasses Swin trained with 3$\times$ schedule (\textbf{49.0} \textit{v.s.} 48.5). For semantic segmentation with UPerNet~\cite{xiao2018unified}, FocalNet base at single-scale outperforms Swin by \textbf{2.4}, and beats Swin at multi-scale (\textbf{50.5} \textit{v.s.} 49.7). Using large FocalNet and Mask2former~\cite{cheng2022masked}, we achieve \textbf{58.5} mIoU for ADE20K semantic segmentation, and \textbf{57.9} PQ for COCO Panoptic Segmentation. Using huge FocalNet and DINO~\cite{zhang2022dino}, we achieved \textbf{64.3} and \textbf{64.4} mAP on COCO \textit{minival} and \textit{test-dev}, respectively, establishing new SoTA on top of much larger attention-based models like Swinv2-G~\cite{liu2022swin} and BEIT-3~\cite{wang2022image}. These encouraging results render \emph{focal modulation is probably what we need for vision}\footnote{Code and models are available at: \url{https://github.com/microsoft/FocalNet}.}.
\end{abstract}

\section{Introduction}
\label{Sec:intro}
Transformers~\cite{vaswani2017attention}, originally proposed for natural language processing (NLP), have become a prevalent architecture in computer vision since the seminal work of Vision Transformer (ViT)~\cite{dosovitskiy2020image}. Its promise has been demonstrated in various vision tasks including image classification~\cite{touvron2020training,wang2021pyramid,wu2021cvt,liu2021swin,zhang2021multi,vaswani2021scaling}, object detection~\cite{carion2020end,zhu2020deformable,zheng2020end,dai2020up}, segmentation~\cite{wang2020max,wang2020end,cheng2021per}, and beyond~\cite{li2021trear,zhao2021tuber,chang2021augmented,chen2021Transformer,wang2021Transformer,li2021sctn}. In Transformers, the self-attention (SA) is arguably the key to its success which enables input-dependent global interactions, in contrast to convolution operation which constrains interactions in a local region with a shared kernel. Despite this advantages, the efficiency of SA has been a concern due to its quadratic complexity over the number of visual tokens, especially for high-resolution inputs. To address this, many works have proposed SA variants through token coarsening~\cite{wang2021pyramid}, window attention~\cite{liu2021swin,vaswani2021scaling,zhang2021multi}, dynamic token selection~\cite{rao2021dynamicvit, yin2022vit, meng2022adavit}, or the hybrid~\cite{yang2021focal,chu2021twins}. Meanwhile, a number of models have been proposed by augmenting SA with (depth-wise) convolutions to capture long-range dependencies with a good awareness of local structures~\cite{wu2021cvt,guo2021cmt,xu2021co,gao2021container,dong2021cswin,lee2022mpvit,chen2022mixformer, ding2022davit}. 

\begin{figure}[t]
	\centering
	\includegraphics[width=0.99\linewidth]{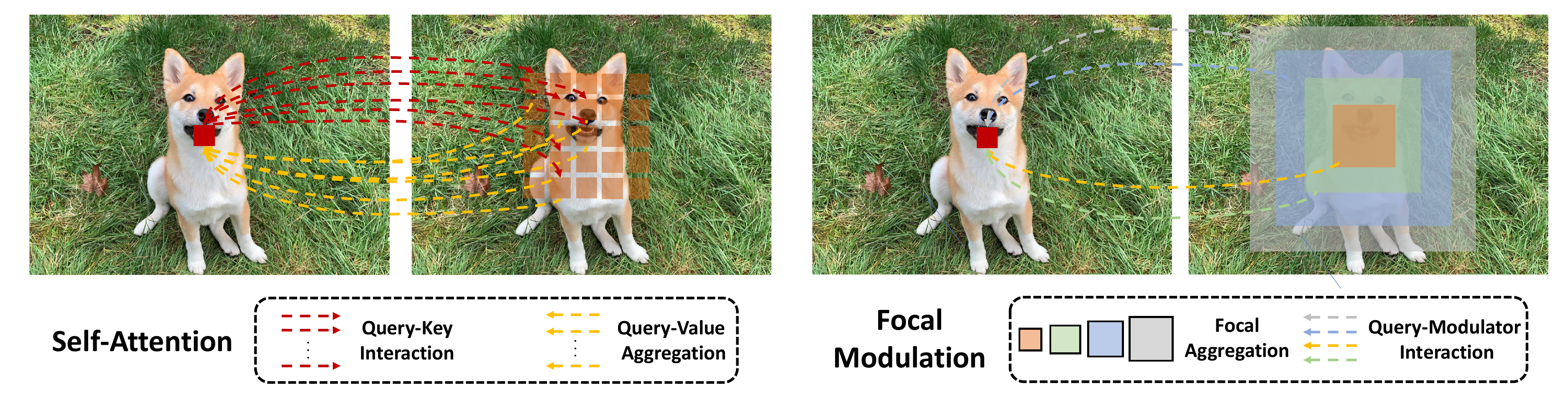}
    \captionsetup{font=footnotesize}
    \caption{Illustrative comparison between (window-wise) Self-Attention (SA)~\cite{vaswani2017attention,dosovitskiy2020image,liu2021swin} and our proposed Focal Modulation. Given the query token \fcolorbox{white}{red}{\rule[0mm]{0pt}{3pt}\rule[0mm]{3pt}{0pt}} and target tokens ( \fcolorbox{white}{orange}{\rule{0pt}{3pt}\rule{3pt}{0pt}},$\cdots$,\fcolorbox{white}{orange}{\rule{0pt}{3pt}\rule{3pt}{0pt}}), SA first performs query-key interactions to compute the attention scores, and then query-value aggregations to capture the context from other tokens. In contrast, Focal Modulation first encodes spatial context at different levels of granularity into \emph{modulators} ( \fcolorbox{white}{orange}{\rule{0pt}{3pt}\rule{3pt}{0pt}} , \fcolorbox{white}{lime}{\rule{0pt}{3pt}\rule{3pt}{0pt}} , \fcolorbox{white}{cyan}{\rule{0pt}{3pt}\rule{3pt}{0pt}}, \fcolorbox{white}{gray}{\rule{0pt}{3pt}\rule{3pt}{0pt}}), which are then adaptively injected into the query token in a query-dependent manner. Clearly, SA requires heavy interaction and aggregation operations, while our Focal Modulation reverses their order and turn both of them light-weight. Figures better viewed in color.
    }
    \vspace{-6mm}
    \label{fig:teaser}
\end{figure}

In this work, we aim at answering the fundamental question: \textit{Is there a better way than SA to model input-dependent long-range interactions?} We start with an analysis on the current advanced designs for SA. In Fig.~\ref{fig:teaser} left side, we show a commonly-used (window-wise) attention between the red query token and its surrounding orange tokens proposed in ViTs~\cite{dosovitskiy2020image} and Swin Transformer~\cite{liu2021swin}. To produce the outputs, SA involve heavy query-key interactions (red arrows) followed by equally heavy query-value aggregations (yellow arrows) between the query and a large number of spatially distributed tokens (context features). However, {is it necessary to undertake such heavy interactions and aggregations?} In this work, we take an alternative way by \emph{first aggregating contexts focally around each query and then adaptively modulating the query with the aggregated context}. As shown in Fig.~\ref{fig:teaser} right side, we can simply apply query-agnostic focal aggregations (\textit{e.g.}, depth-wise convolution) to generate summarized tokens at different levels of granularity. Afterwards, these summarized tokens are adaptively aggregated into a \textit{modulator}, which is finally injected into the query. 
\begin{wrapfigure}{r}{0.45\textwidth}
  \begin{center}
  \vspace{-7mm}
   \hspace{-10mm}
    \includegraphics[width=0.5\textwidth]{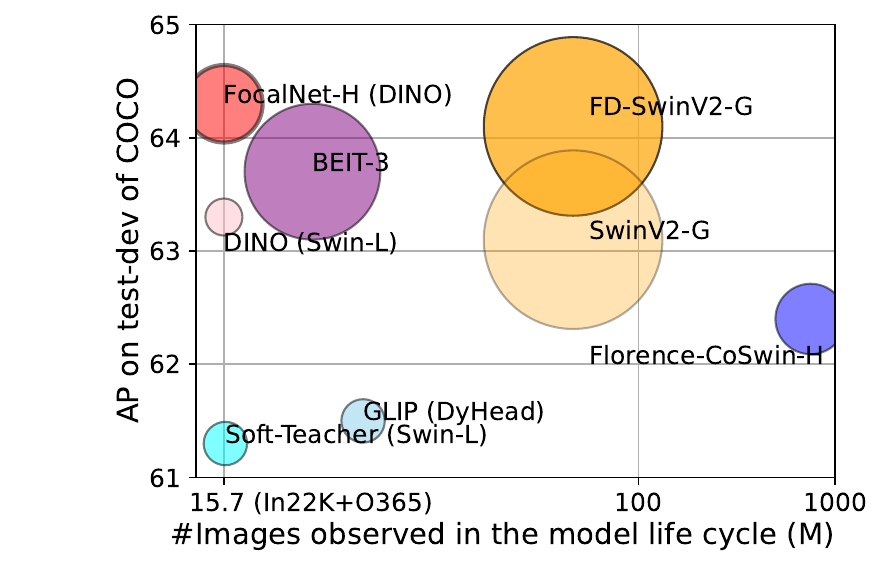}
  \end{center}
    \vspace{-4mm}
    \captionsetup{font=footnotesize}    
    \caption{Comparison with SoTA on COCO object detection. Circle size indicates the model size. }
    \vspace{-4mm}
    \label{fig:coco_sota}
\end{wrapfigure}
This alteration still enables input-dependent token interaction, but significantly eases the process by decoupling the aggregation from individual queries, hence making the interactions light-weight upon merely a couple of features. Our method is inspired by focal attention~\cite{yang2021focal} which performs multiple levels of aggregation to capture fine- and coarse-grained visual contexts. However, our method extracts at each query position the modulator and exploits a much simpler way for the query-modulator interaction. We call this new mechanism \textit{Focal Modulation}, with which we replace SA to build an attention-free architecture, \textit{Focal Modulation Network}, or \emph{FocalNet} in short.

Finally, extensive experiments on image classification, object detection and segmentation, show our FocalNets consistently and significantly outperform the SoTA SA counterparts with comparable costs. Notably, our FocalNet achieves \textbf{82.3}\% and \textbf{83.9}\% top-1 accuracy using tiny and base model size, but with comparable and doubled throughput than Swin and Focal Transformer, respectively. When pretrained on ImageNet-22K with $224^2$ resolution, our FocalNets achieve \textbf{86.5}\% and \textbf{87.3}\% in $224^2$ and $384^2$ resolution, respectively, which are comparable or better than Swin at similar cost. The advantage is particularly significant when transferred to dense prediction tasks. For object detection on COCO~\cite{lin2014microsoft}, our FocalNets with tiny and base model size achieve \textbf{46.1} and \textbf{49.0} box mAP on Mask R-CNN 1$\times$, surpassing Swin with 3$\times$ schedule ({46.0} and {48.5} box mAP). For semantic segmentation on ADE20k~\cite{zhou2017scene}, our FocalNet with base model size achieves \textbf{50.5} mIoU at single-scale evaluation, outperforming Swin at multi-scale evaluation ({49.7} mIoU). Using the pretrained large FocalNet, we achieve \textbf{58.5} mIoU for ADE20K semantic segmentation, and \textbf{57.9} PQ for COCO Panoptic Segmentation based on Mask2former~\cite{cheng2021masked}. Using huge FocalNet and DINO~\cite{zhang2022dino}, we achieved \textbf{64.3} and \textbf{64.4} mAP on COCO \textit{minival} and \textit{test-dev}, respectively, establishing new SoTA on COCO over much larger attention-based models like Swinv2-G~\cite{liu2022swin} and BEIT-3~\cite{wang2022image}. Please find the visual comparison in Figure~\ref{fig:coco_sota}, and details in the experiments. Finally, we apply our Focal Modulation in monolithic layout as ViTs and clearly demonstrate its superiority across different model sizes.

\section{Related Work}
\label{Sec:RelatedWork}
\textbf{Self-attentions}. 
Transformer~\cite{vaswani2017attention} is first introduced to vision in Vision Transformer (ViT)~\cite{dosovitskiy2020image} by splitting an image into a sequence of visual tokens. The self-attention (SA) strategy in ViTs has demonstrated superior performance to modern convolutional neural networks (ConvNets) such as ResNet~\cite{he2016deep} when trained with optimized recipes~\cite{dosovitskiy2020image,touvron2020training}. Afterwards, multi-scale architectures~\cite{chen2021crossvit,wang2021pyramid,xu2021co}, light-weight convolution layers~\cite{wu2021cvt,guo2021cmt,li2021localvit}, local self-attention mechanisms~\cite{liu2021swin,zhang2021multi,chu2021twins, yang2021focal} and learnable attention weights~\cite{yuan2021volo} have been proposed to boost the performance and support high-resolution input. More comprehensive surveys are covered in~\cite{khan2021transformers,han2020survey,khan2021transformers}. Our focal modulation significantly differs from SA by first aggregating the contexts from different levels of granularity and then modulating individual query tokens, rendering an attention-free mechanism for token interactions. For context aggregation, our method is inspired by focal attention proposed in~\cite{yang2021focal}. However, the context aggregation for focal modulation is performed at each query location instead of target locations, followed by a modulation rather than an attention. These differences in mechanism lead to significant improvement of efficiency and performance. Another closely related work is Poolformer~\cite{yu2021metaformer} which uses a pooling to summarize the local context and a simple subtraction to adjust the individual inputs. Despite decent efficiency, it lags behind popular vision transformers like Swin on performance. As we will show, capturing local structures at different levels is essential. 

\textbf{MLP architectures}. 
%
Visual MLPs can be categorized into two groups: $(i)$ Global-mixing MLPs, such as MLP-Mixer~\cite{mlp-mixer} and ResMLP~\cite{resmlp}, perform global communication among visual tokens through spatial-wise projections augmented by various techniques, such as gating, routing, and Fourier transforms~\cite{gmlp,lou2021sparse,tang2021sparse,tang2021image}.
$(ii)$ Local-mixing MLPs sample nearby tokens for interactions, using spatial shifting, permutation, and pseudo-kernel mixing~\cite{yu2021s2,hou2021vision,lian2021mlp,chen2021cyclemlp,guo2021hire}. Recently, Mix-Shift-MLP~\cite{zheng2022mixing} exploits both local and global interactions with MLPs, in a similar spirit of focal attention~\cite{yang2021focal}. Both MLP architectures and our focal modulation network are attention-free. However, focal modulation  with multi-level context aggregation naturally captures the structures in both short- and long-range, and thus achieves much better accuracy-efficiency trade-off.


\textbf{Convolutions}. ConvNets have been the primary driver of the renaissance of deep neural networks in computer vision.
The field has evolved rapidly since the emerge of VGG~\cite{simonyan2014very}, InceptionNet~\cite{szegedy2015going} and ResNet~\cite{he2016deep}. Representative works that focus on the efficiency of ConvNets are MobileNet~\cite{howard2017mobilenets}, ShuffleNet~\cite{zhang2018shufflenet} and EfficientNet~\cite{tan2019efficientnet}. Another line of works aimed at integrating global context to compensate ConvNets such as SE-Net~\cite{hu2018squeeze}, Non-local Network~\cite{wang2018non}, GCNet~\cite{cao2019gcnet}, LR-Net~\cite{hu2019local} and C3Net~\cite{yang2019cross}, \textit{etc}. Introducing dynamic operation is another way to augment ConvNets as demonstrated in Involution~\cite{li2021involution} and DyConv~\cite{chen2020dynamic}. 
Recently, ConvNets strike back from two aspects: 
$(i)$ convolution layers are integrated to SA and bring significant gains~\cite{wu2021cvt,guo2021cmt,li2021localvit,gao2021container} or the vice versa~\cite{touvron2021augmenting}; $(ii)$ ResNets have closed the gap to ViTs using similar data augmentation and regularization strategies~\cite{wightman2021resnet}, and replacing SA with (dynamic) depth-wise convolution~\cite{han2021demystifying, liu2022convnet} can also slightly surpass Swin. Our focal modulation network also exploits depth-wise convolution as the micro-architecture but goes beyond by introducing a multi-level context aggregation and input-dependent modulation.
We will show this new module significantly outperforms raw convolution networks.






\section{Focal Modulation Network}
\label{Sec:Method}
\subsection{From Self-Attention to Focal Modulation}
\vspace{-2pt}
Given a visual feature map $\Xmat \in \R^{H\times W \times C}$ as input, a generic encoding process generates for each visual token (query) $\xv_i \in \R^{C}$ a feature representation $\yv_i \in \R^{C}$ via the interaction $\mathcal{T}$ with its surroundings $\Xmat$ (\eg neighboring tokens) and aggregation $\mathcal{M}$ over the contexts. 

\textbf{Self-attention}. The self-attention modules use a late aggregation procedure formulated as  
\begin{equation}
\small
    \yv_i = \mathcal{M}_1 ( \mathcal{T}_1 (\xv_i, \Xmat),  \Xmat ),  \hspace{-0mm}
    \label{Eq:late_agg}
\end{equation}
where the aggregation $\mathcal{M}_1$  over the contexts $\Xmat$ is performed after the attention scores between query and target are computed via interaction $\mathcal{T}_1$.

\textbf{Focal modulation}. In contrast, Focal Modulation generates refined representation $\yv_i$ using an early aggregation procedure formulated as
\begin{equation}
\small
    \yv_i = \mathcal{T}_2 ( \mathcal{M}_2 (i, \Xmat),  \xv_i ), \hspace{0mm}
    \label{Eq:visual_modulation}
\end{equation}
where the context features are first aggregated using $\mathcal{M}_2$ at each location $i$, then the query interacts with the aggregated feature based on $\mathcal{T}_2$ to form $\yv_i$. 
%
%

Comparing Eq.~\eqref{Eq:late_agg} and Eq.~\eqref{Eq:visual_modulation}, we see that
$(i)$
the context aggregation of Focal Modulation $\mathcal{M}_2$ amortizes the computation of contexts via a shared operator (\textit{e.g.}, depth-wise convolution), while $\mathcal{M}_1$ in SA is more computationally expensive as it requires summing over non-shareable attention scores for different queries;  
$(ii)$ the interaction $\mathcal{T}_2$ is a lightweight operator between a token and its context, while $\mathcal{T}_1$ involves computing token-to-token attention scores, which has quadratic complexity. 

\begin{figure}[t]
	\centering
	\includegraphics[width=0.96\linewidth]{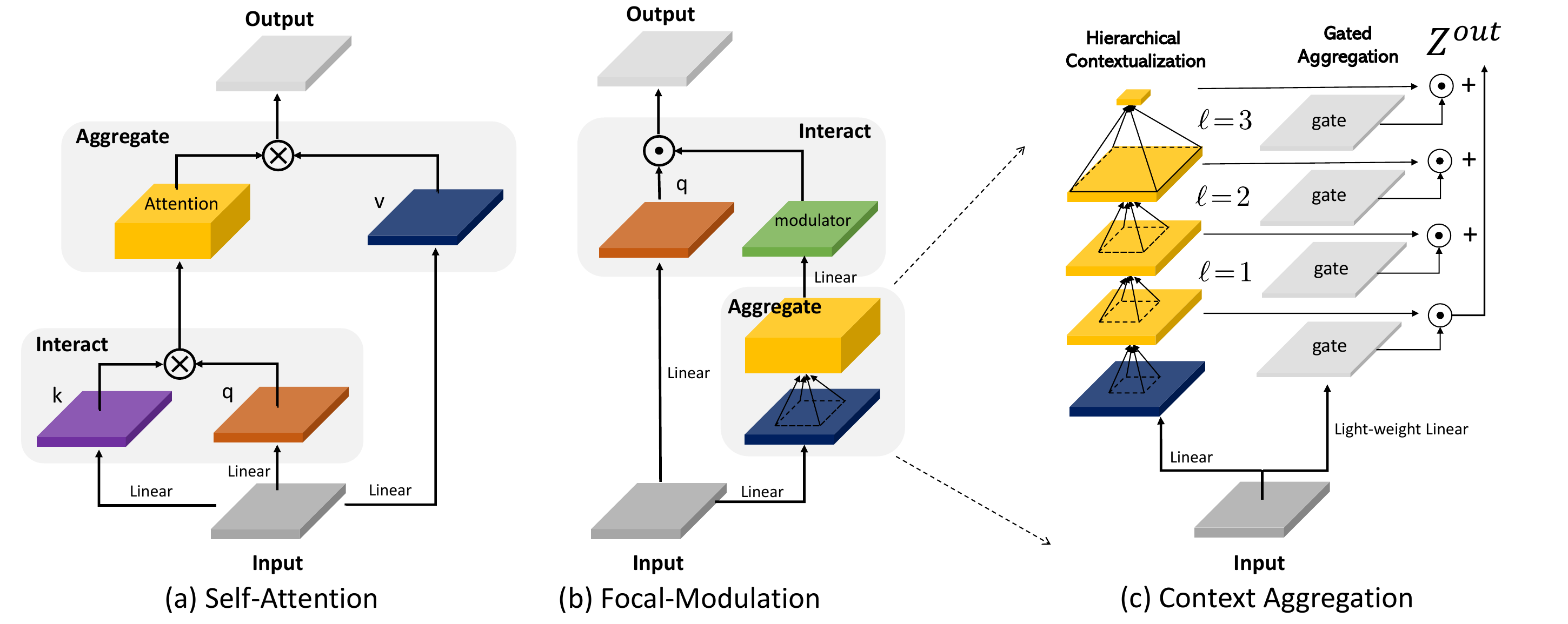}
\captionsetup{font=footnotesize}
    \caption{Left: Comparing SA (a) and Focal Modulation (b) side by side. Right: Detailed illustration of context aggregation in Focal Modulation (c).}
    \label{fig:model}
\vspace{-3mm}
\end{figure}

Based on Eq.~\eqref{Eq:visual_modulation}, we instantiate our Focal Modulation to
\begin{equation}
    \yv_i = q(\xv_i) \odot m (i, \Xmat),  \hspace{-5mm}
    \label{Eq:focal_modulation}
\end{equation}
where $q(\cdot)$ is a query projection function and 
$\odot$ is the element-wise multiplication. $m(\cdot)$ is a context aggregation function, whose output is called \emph{modulator}. Fig.~\ref{fig:model}(a) and (b) compare Self-Attention and Focal Modulation. 
The proposed Focal Modulation has the following favorable properties:
\begin{itemize}[leftmargin=*]
  \setlength\itemsep{0em}
    \item \textbf{Translation invariance}. Since $q(\cdot)$ and $m(\cdot)$ are always centered at the query token $i$ and no positional embedding is used, the modulation is invariant to translation of input feature map $\Xmat$.
    \item \textbf{Explicit input-dependency}. The modulator is computed via $m(\cdot)$ by aggregating the local features around target location $i$, hence our Focal Modulation is explicitly input-dependent.
    \item \textbf{Spatial- and channel-specific}. The target location $i$ as a pointer for $m(\cdot)$  enables spatial-specific modulation. The element-wise multiplication enables channel-specific modulation. 
    \item \textbf{Decoupled feature granularity}. $q(\cdot)$ preserve the finest information for individual tokens, while $m(\cdot)$ extracts the coarser context. They are decoupled but combined through modulation. 
\end{itemize}


In what follows, we describe in detail the implementation of $m(\cdot)$ 
 in Eq.~\eqref{Eq:focal_modulation}.

\subsection{Context Aggregation via $m(\cdot)$}
\label{sec:focal_conv}
It has been proved that both short- and long-range contexts are important for visual modeling~\cite{yang2021focal,dong2021cswin,liu2022convnet}. However, a single aggregation with larger receptive field is not only computationally expensive in time and memory, 
but also undermines the local fine-grained structures which are particularly useful for dense prediction tasks. 
Inspired by~\cite{yang2021focal}, we propose a multi-scale hierarchical context aggregation. As depicted in Fig.~\ref{fig:model}~(c), the aggregation procedure consists of two steps: {\it hierarchical contextualization} to extract contexts from local to global ranges at different levels of granularity and {\it gated aggregation} to condense all context features at different granularity levels into the modulator.

\textbf{Step 1: Hierarchical Contextualization.}

Given input feature map $\Xmat$, we first project it into a new feature space with a linear layer $\Zmat^0=f_z(\Xmat) \in \R^{H \times W \times C}$.
Then, a hierarchical presentation of contexts is obtained using a stack of $L$ depth-wise convolutions.
At focal level $\ell \in \{1,...,L\}$, the output  $\Zmat^{\ell}$ is derived by:
\begin{equation}
\small
\Zmat^{\ell} = f_a^{\ell}(\Zmat^{\ell-1}) \triangleq \GeLU( \DWConv( \Zmat^{\ell-1} )) \in \R^{H \times W \times C},
\label{eq:hier_context}
\end{equation}
where $f_{a}^\ell$ is the contextualization function at the $\ell$-th level, implemented via a depth-wise convolution $\DWConv$ with kernel size $k^\ell$ followed by a $\GeLU$ activation function~\cite{hendrycks2016gaussian}. 
The use of depth-wise convolution for hierarchical contextualization of Eq.~\eqref{eq:hier_context} is motivated by its desirable properties. Compared to pooling~\cite{yu2021metaformer,hu2018squeeze}, depth-wise convolution is learnable and structure-aware. In contrast to regular convolution, it is channel-wise and thus computationally much cheaper. 

Hierarchical contextualization of Eq.~\eqref{eq:hier_context} generates $L$ levels of feature maps. At level $\ell$, the effective receptive field is $r^\ell = 1 + \sum_{i=1}^\ell (k^\ell-1)$, which is much larger than the kernel size $k^\ell$. 
To capture global context of the whole input, which could be high-resolution, 
we apply a global average pooling on the $L$-th level feature map $\Zmat^{L+1} = \AvgPool(\Zmat^L)$. Thus, we obtain in total $(L+1)$ feature maps $\{\Zmat^\ell\}_{\ell=1}^{L+1}$, which collectively capture short- and long-range contexts at different levels of granularity. 


\textbf{Step 2: Gated Aggregation.}
\begin{figure}[t]
	\begin{minipage}{0.52\linewidth}
	\centering
    \includegraphics[width=1.0\linewidth]{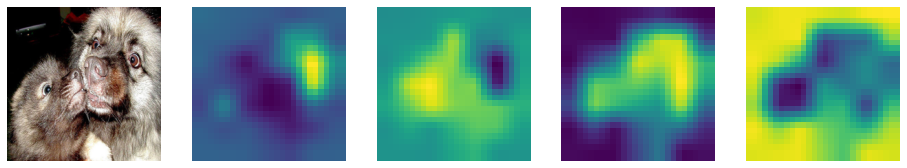}
    \includegraphics[width=1.0\linewidth]{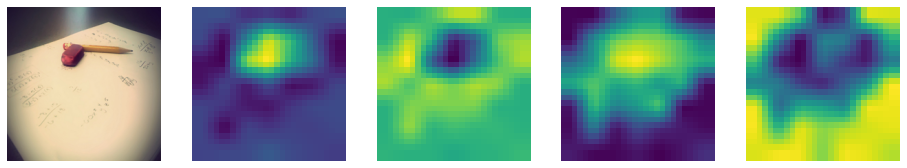}
	\vspace{-5mm}
    \captionsetup{font=footnotesize}
    \caption{Visualization of gating values ${\Gmat}$ in Eq.~\eqref{eq:gated_agg} at last layer of our FocalNet ($L=3$) pretrained on ImageNet-1K. The columns from left to right are input images, gating maps at focal level 1,2,3 and global level.}
    \label{fig:gating_vis_main}	
	\end{minipage}	
	\quad
	\begin{minipage}{0.44\linewidth}
	\centering
	\includegraphics[width=1.0\linewidth]{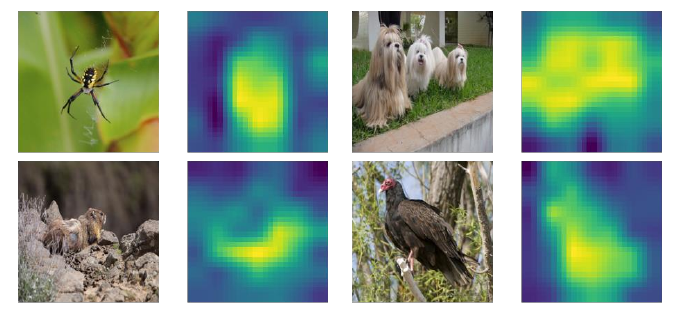}	
	\vspace{-6mm}
    \captionsetup{font=footnotesize}
    \caption{Visualization of modulator values (corresponding to the right side of $\odot$ in Eq.~\eqref{Eq:FocalModulation}) at the last layer in FocalNet. The original modulator map is upsampled for display.}
    \label{fig:modulation}	
	\end{minipage}	
	\vspace{-1mm}
\end{figure}

In this step, the $(L+1)$ feature maps obtained via hierarchical contextualization are condensed into a {modulator}.
In an image, the relation between a visual token (query) and its surrounding contexts often depends on the content itself. For example, the model might rely on local fine-grained features for encoding the queries of salient visual objects, but mainly global coarse-grained features for the queries of background scenes.
Based on this intuition, we use a gating mechanism to control how much to aggregate from different levels for each query. 
Specifically, we use a linear layer to obtain a spatial- and level-aware gating weights $ \Gmat = f_g(\Xmat) \in \R^{H \times W \times (L+1)}$. Then, we perform a weighted sum through an element-wise multiplication to obtain a single feature map $\Zmat^{out}$ which has the same size as the input $\Xmat$,
\begin{equation}
\small
    \Zmat^{out} = \sum_{\ell=1}^{L+1} \Gmat^\ell \odot \Zmat^\ell \in \R^{H \times W \times C}
\label{eq:gated_agg}
\end{equation}
where $\Gmat^{\ell} \in \R^{H \times W \times 1}$ is a slice of $\Gmat$ for the level $\ell$. When visualizing these gating maps in Fig.~\ref{fig:gating_vis_main}, we surprisingly find our FocalNet indeed learns gathering the context from different focal levels adaptively as we expect. As we can see, for a token on a small object, it focuses more on the fine-grained local structure at low focal level, while a token in a uniform background needs to be aware of much larger contexts from higher levels. 
Until now, all the aggregation is spatial. To enable the communication across different channels, we use another linear layer $h(.)$ to obtain the modulator map $\Mmat = h(\Zmat^{out}) \in \R^{H \times W \times C}$. In Fig.~\ref{fig:modulation}, we visualize the magnitude of modulator $\Mmat$ at the last layer of our FocalNet. Interestingly, the modulators automatically pay more attention to the objects inducing the category, which implies a simple way of interpreting FocalNets.

\paragraph{Focal Modulation.} Given the implementation of $m(\cdot)$ as described above, Focal Modulation of Eq.\eqref{Eq:focal_modulation} can be rewritten at the token level as
\begin{equation}
    \yv_i = q(\xv_i) \odot h(\sum_{\ell=1}^{L+1} \gv^\ell_i \cdot \zv^\ell_i)
    \label{Eq:FocalModulation}
\end{equation}
where $\gv_i^\ell$ and $\zv_i^\ell$ are the gating value and visual feature at location $i$ of $\Gmat^\ell$ and $\Zmat^\ell$, respectively. We summarize the proposed Focal Modulation in Pytorch-style pseudo code in Algorithm~\ref{alg:focalnet}, which is implemented with a few depth-wise convolution and linear layers.

\begin{algorithm}[t]
  \caption{Pseudo code for Focal Modulation.}
  \label{alg:focalnet}
  \scriptsize

  \Comment{Input/output shape: (B, H, W, C); Batchsize B; Feature map height H, width W, dim C}
  \Comment{Focal levels: L;  Conv kernel size at level $\ell$: k$^{\ell}$}

  \Function{init( )}{
     \var{pj\_in, pj\_cxt = Linear(C, 2*C + (L+1)), Conv2d(C, C, 1) }\; 
     \var{hc\_layers = [Sequential(Conv2d(C, C, k$^{\ell}$, groups=C), GeLU())\forinline{for} $\ell$ in range(L)]}\;
     \var{pj\_out = Sequential(Linear(C, C),  Dropout()) }\;
  }

  \Function{forward(\var{x}, \var{m=0})}{ 
    \var{x = \FuncCall{pj\_in}{x}.\FuncCall{permute}{0, 3, 1, 2} } \;
    \var{q, z, gate = split(x, (C, C, L+1), 1)} \;

     \For{$\ell$  in range(L)}{
    \var{ z = hc\_layers[$\ell$](z) }
         \CommentSty{\hspace{11mm}\# Eq.\eqref{eq:hier_context}, hierarchical contextualization}

     \var{ m = m +  z * gate[:, $\ell$:$\ell$+1] }
     \CommentSty{~\# Eq.\eqref{eq:gated_agg}, gated aggregation}
     }
     \var{m = m + GeLU(z.mean(dim=(2,3))) * gate[:,L:]}
     
    \var{x = q * pj\_cxt(m) }           
    \CommentSty{\hspace{20mm}\# Eq.\eqref{Eq:FocalModulation}, Focal Modulation}    \; 
    \Return { pj\_out( x.permute(0, 2, 3, 1) ) }
    }
     
\end{algorithm}

\subsection{Relation to Other Architecture Designs} 
\label{sec:discussions}
Based on the formula in Eq.~\eqref{Eq:FocalModulation}, we build the connections between our Focal Modulation and other relevant architecture designs beyond Self-Attention.


\textbf{Depth-wise Convolution} has been used to augment the local structural modeling for SA~\cite{wu2021cvt,dong2021cswin,guo2021cmt} or purely to enable efficient long-range interactions~\cite{howard2017mobilenets,han2021demystifying,liu2022convnet}. Our Focal Modulation also employs depth-wise convolution as one of the building blocks. However, instead of using its response as the output directly, our Focal Modulation uses depth-wise convolution to capture the hierarchical contexts, which are then converted into modulator to modulate each query. As we will show in our experiments, these three components as a whole contribute the final decent performance.

\textbf{Squeeze-and-Excitation (SE)} was proposed in~\cite{hu2018squeeze} prior to the emerge of vision transformers. It exploits a global average pooling to squeeze the context globally, and then a multi-layer perception (MLP) followed by a $\Sigmoid$ to obtain the excitation scalar for each channel. SE can be considered as a special case of Focal Modulation. Setting $L=0$ in Eq.~\eqref{Eq:FocalModulation}, Focal Modulation degrades to $q(\xv_i) \odot h(f_g(\xv_i) \cdot \AvgPool(f_z(\Xmat)))$ which resembles SE. In our experiments, we study this variant and find that a global context is far insufficient for visual modeling.

\textbf{PoolFormer} was recently introduced in~\cite{yu2021metaformer}, and draw many attentions due to its simplicity. It uses average pooling to extract the context locally in a sliding-window, and then adjust the query tokens using an element-wise subtraction. It shares similar spirit to SE-Net, but uses local context instead of global ones, and subtraction instead of multiplication. Putting it and Focal Modulation side-by-side, we can find both of them extract the local context and enable the query-context interaction but in different ways (Pooling \textit{v.s.} Convolution, Subtraction \textit{v.s.} Modulation). 

\subsection{Complexity} 
In Focal Modulation as Eq.~\eqref{Eq:FocalModulation}, there are mainly three linear projections $q(\cdot)$, $h(\cdot)$, and $f_{z}(\cdot)$ for $\Zmat^0$. Besides, it requires a lightweight linear function $f_g(\cdot)$ for gating and $L$ depth-wise convolution $f_a^{\{1,...,L\}}$ for hierarchical contextualization. Therefore, the overall number of learnable parameters is $3C^2 + C(L+1) + C \sum_\ell (k^\ell)^2$. Since $L$ and $(k^\ell)^2$ are typically much smaller than $C$, the model size is mainly determined by the first term as we will show in Sec.~\ref{Sec:Experiment}. Regarding the time complexity, besides the linear projections and the depth-wise convolution layers, the element-wise multiplications introduce $\mathcal{O}(C(L+2))$ for each visual token. Hence, the total complexity for a feature map is $\mathcal{O}(HW \times (3C^2 + C(2L+3) + C\sum_\ell (k^\ell)^2))$. For comparison, a window-wise attention in Swin Transformer with window size $w$ is $\mathcal{O}(HW \times (3C^2 + 2Cw^2))$, while a vanilla self-attention  in ViTs takes $\mathcal{O}((HW)^2C + HW \times (3C^2))$.

\subsection{Network Architectures}
%
We use the same stage layouts and hidden dimensions as in Swin~\cite{liu2021swin} and Focal Transformers~\cite{yang2021focal}, but replace the SA modules with the Focal Modulation modules. We thus construct a series of Focal Modulation Network (FocalNet) variants. In FocalNets, we only need to specify the number of focal levels ($L$) and the kernel size ($k^\ell$) at each level. For simplicity, we gradually increase the kernel size by 2 from lower focal levels to higher ones, \textit{i.e.}, $k^\ell = k^{\ell-1}+2$. To match the complexities of Swin and Focal Transformers, we design a small receptive field (SRF) and a large receptive field (LRF) version for each of the four layouts by using 2 and 3 focal levels, respectively. We use non-overlapping convolution layers for patch embedding at the beginning (kernel size=$4\times 4$, stride=4) and between two stages (kernel size=$2 \times 2$, stride=2), respectively.

\section{Experiment}
\vspace{-2mm}
\label{Sec:Experiment}

\subsection{Image Classification}
We compare different methods on ImageNet-1K classification~\cite{deng2009imagenet}.
Following the recipes in~\cite{touvron2020training,liu2021swin,yang2021focal}, we train FocalNet-T, FocalNet-S and FocalNet-B with ImageNet-1K training set and report Top-1 accuracy (\%) on the validation set. Training details are described in the appendix. 


To verify the effectiveness of FocalNet, we compare it with three groups of methods based on ConvNets, Transformers and MLPs. The results are reported in Table~\ref{tab:image_classification}. We see that FocalNets outperform the conventional CNNs (\textit{e.g.}, ResNet~\cite{he2016deep} and the augmented version~\cite{wightman2021resnet}), MLP architectures such as MLP-Mixer~\cite{tolstikhin2021mlp} and gMLP~\cite{liu2021pay}, and Transformer architectures DeiT~\cite{touvron2020training} and PVT~\cite{wang2021pyramid}. 
In particular, we compare FocalNets against Swin and Focal Transformers which use the same architecture to verify FocalNet's stand-alone effectiveness at the bottom part. We see that FocalNets with small receptive fields (SRF) achieve consistently better performance than Swin Transformer but with similar model size, FLOPs and throughput. 
For example, the tiny FocalNet improves Top-1 accuracy by $0.9\%$ over Swin-Tiny. To compare with Focal Transformers (FocalAtt), we change to large receptive fields (LRF) though it is still much smaller than the one used in FocalAtt. Focal modulation outperforms the strong and sophisticatedly designed focal attention across all model sizes. More importantly, its run-time speed is much higher than FocalAtt by getting rid of many time-consuming operations like rolling and unfolding. 

\textbf{Model augmentation.} 
We investigate whether some commonly used techniques for vision transformers can also improve our FocalNets. First, we study the effect of using overlapped patch embedding for downsampling~\cite{guo2021cmt}. Following~\cite{wu2021cvt}, we change the kernel size and stride from $(4, 4)$ to $(7, 4)$ for patch embedding at the beginning, and $(2,2)$ to $(3,2)$ for later stages. The comparisons are reported in Table~\ref{tab:overlapped_patchembed}. Overlapped patch embedding improves the performance for models of all sizes, with slightly increased computational complexity and time cost. Second, we make our FocalNets deeper but thinner as in~\cite{dong2021cswin,zhou2021refiner}. 
In Table~\ref{tab:deeper_network}, we change the depth layout of our FocalNet-T from 2-2-6-2 to 3-3-16-3, and FocalNet-S/B from 2-2-18-2 to 4-4-28-4. Meanwhile, the hidden dimension at first stage is reduced from 96, 128 to 64, 96, respectively. These changes lead to smaller model sizes and fewer FLOPs, but higher time cost due to the increased number of sequential blocks. It turns out that going deeper improves the performance of FocalNets significantly. These results demonstrate that the commonly used model augmentation techniques developed for vision transformers can be easily adopted to improve the performance of FocalNets. 

\begin{table}[t!]
\begin{minipage}{0.48\linewidth}
\footnotesize
\centering
\setlength{\tabcolsep}{1.8pt}
\resizebox{0.96\linewidth}{!}{
    \begin{tabular}{l|cccc}
    \toprule
    Model & \makecell{\#Params. \\ (M)} & \makecell{FLOPs \\ (G)} & \makecell{Throughput \\ (imgs/s)} & \makecell{Top-1 \\ (\%)} \\
    \midrule
    ResNet-50~\cite{he2016deep} & 25.0 & 4.1 & 1294 &  76.2 \\
    ResNet-101~\cite{he2016deep}  & 45.0 & 7.9 & 745 & 77.4 \\
    ResNet-152~\cite{he2016deep}  & 60.0 & 11.0 & 522 & 78.3 \\
    ResNet-50-SB~\cite{wightman2021resnet} & 25.0 & 4.1 & 1294 & 79.8 \\
    ResNet-101-SB~\cite{wightman2021resnet} & 45.0 & 7.9 & 745 & 81.3 \\
    ResNet-152-SB~\cite{wightman2021resnet} & 60.0 & 11.6 & 522 & 81.8 \\
    DW-Net-T~\cite{han2021demystifying} & 24.2 & 3.8 & 1030 & 81.2 \\
    DW-Net-B~\cite{han2021demystifying} & 74.3 & 12.9 & 370 & 83.2 \\
    \midrule
    Mixer-B/16~\cite{tolstikhin2021mlp} & 59.9 & 12.7 & 455 & 76.4 \\
    gMLP-S~\cite{liu2021pay} & 19.5 & 4.5 & 785 & 79.6 \\
    gMLP-B~\cite{liu2021pay} & 73.4 & 15.8 & 301 & 81.6 \\
    ResMLP-S24~\cite{resmlp} & 30.0 & 6.0  &  871 & 79.4 \\
    ResMLP-B24~\cite{resmlp} & 129.1 & 23.0 & 61  & 81.0 \\
    \midrule    
    DeiT-Small/16~\cite{touvron2020training} & 22.1 & 4.6 & 939 & 79.9 \\
    DeiT-Base/16~\cite{touvron2020training}& 86.6 & 17.5 & 291 & 81.8 \\
    PVT-Small~\cite{wang2021pyramid}  & 24.5 & 3.8 & 794 & 79.8 \\
    PVT-Medium~\cite{wang2021pyramid} & 44.2 & 6.7 & 517 & 81.2 \\
    PVT-Large~\cite{wang2021pyramid} & 61.4 & 9.8 & 352 & 81.7 \\
    PoolFormer-m36~\cite{yu2021metaformer} & 56.2 & 8.8 & 463 & 82.1 \\ 
    PoolFormer-m48~\cite{yu2021metaformer} & 73.5 & 11.6 & 347 & 82.5 \\
    \midrule
    Swin-Tiny~\cite{liu2021swin}   & 28.3 & 4.5 & 760 & 81.2 \\
    \rowcolor{Gray}
    FocalNet-T~(SRF)               & 28.4 & 4.4 & 743 & \textbf{82.1} \\
    Swin-Small~\cite{liu2021swin}    &  49.6 & 8.7 & 435 & 83.1 \\    
    \rowcolor{Gray}
    FocalNet-S~(SRF)               & 49.9 & 8.6 & 434 & \textbf{83.4} \\
    Swin-Base~\cite{liu2021swin}  & 87.8 & 15.4 & 291 & 83.5 \\  
    \rowcolor{Gray}
    FocalNet-B~(SRF) & 88.1 & 15.3 & 280 & \textbf{83.7} \\
    \hline
    FocalAtt-Tiny~\cite{yang2021focal} & 28.9 & 4.9 & 319 & {82.2} \\
    \rowcolor{Gray}
    FocalNet-T~(LRF)    & 28.6 & 4.5 & 696 & \textbf{82.3}  \\
    FocalAtt-Small & 51.1 & 9.4 & 192 & \textbf{83.5} \\
    \rowcolor{Gray}
    FocalNet-S~(LRF) & 50.3 & 8.7 & 406 & \textbf{83.5} \\   
    FocalAtt-Base~\cite{yang2021focal}  & 89.8 & 16.4 & 138 & 83.8 \\
    \rowcolor{Gray}
    FocalNet-B~(LRF)  &  88.7 & 15.4 & 269 & \textbf{83.9} \\
     \bottomrule
    \end{tabular}
    
    }
    \captionsetup{font=footnotesize}  
    \caption{ImageNet-1K classification comparison.}
    \label{tab:image_classification}
\vspace{-3mm}
\end{minipage}
\begin{minipage}{0.48\linewidth}
\begin{minipage}{1.0\linewidth}
\footnotesize
\centering
\setlength{\tabcolsep}{1.8pt}
\resizebox{0.97\linewidth}{!}{
    \begin{tabular}{l|ccccc}
    \toprule
    Model & \makecell{Overlapped \\ PatchEmbed} & \makecell{\#Params. \\ (M)} & \makecell{FLOPs \\ (G)} & \makecell{Throughput \\ (imgs/s)} & \makecell{Top-1 \\ (\%)} \\
    \midrule
    FocalNet-T~(SRF)  & & 28.4 & 4.4 & 743 & 82.1 \\
    \rowcolor{Gray}
    FocalNet-T~(SRF)  & \checkmark & 30.4 & 4.4 & 730 & \textbf{82.4} \\
    FocalNet-S~(SRF) & & 49.9 & 8.6 & 434 & 83.4 \\
    \rowcolor{Gray}
    FocalNet-S~(SRF) & \checkmark & 51.8 & 8.6 & 424 & \textbf{83.4} \\
    FocalNet-B~(SRF)  & & 88.1 & 15.3 & 286 & 83.7 \\
    \rowcolor{Gray}
    FocalNet-B~(SRF)  & \checkmark & 91.6 & 15.3 & 278 & \textbf{84.0}  \\    
     \bottomrule
    \end{tabular}
    }
    \captionsetup{font=footnotesize}  
    \vspace{1mm}
    \caption{Effect of overlapped patch embedding.}
    \label{tab:overlapped_patchembed}
\end{minipage}
\begin{minipage}{1.0\linewidth}
\footnotesize
\centering
\setlength{\tabcolsep}{1.8pt}
\resizebox{0.97\linewidth}{!}{
    \begin{tabular}{l|cccccc}
    \toprule
    Model & \makecell{Depth} & Dim. & \makecell{\#Params.} & \makecell{FLOPs} & \makecell{Throughput} & \makecell{Top-1} \\
    \midrule
    FocalNet-T~(SRF)  & 2-2-6-2 & 96 & 28.4 & 4.4 & 743 & 82.1 \\
    \rowcolor{Gray}
    FocalNet-T~(SRF)  & 3-3-16-3 & 64 & 25.1 & 4.0 & 663 & \textbf{82.7} \\
    FocalNet-S~(SRF) & 2-2-18-2 & 96 & 49.9 & 8.6 & 434 & 83.4 \\
    \rowcolor{Gray}
    FocalNet-S~(SRF) & 4-4-28-4 & 64 & 38.2 & 6.4 & 440 & \textbf{83.5} \\
    FocalNet-B~(SRF)  & 2-2-18-2 & 128 & 88.1 & 15.3 & 280 & 83.7 \\
    \rowcolor{Gray}
    FocalNet-B~(SRF)  & 4-4-28-4 & 96 & 85.1 & 14.3 & 247 & \textbf{84.1} \\    
     \bottomrule
    \end{tabular}
    }
    \vspace{1mm}
    \captionsetup{font=footnotesize}
    \caption{Effect of deeper and thinner networks.}
    \label{tab:deeper_network}
\end{minipage}
\begin{minipage}{1.0\linewidth}
\footnotesize
\centering
\setlength{\tabcolsep}{1.8pt}
\resizebox{0.97\linewidth}{!}{
    \begin{tabular}{l|ccccc}
    \toprule
    Model & Img. Size & {\#Params} & {FLOPs} & {Throughput} & {Top-1} \\
    \midrule
 ResNet-101x3~\cite{he2016deep} & 384$^2$ & 388.0 & 204.6 & - & 84.4 \\
 ResNet-152x4~\cite{he2016deep} & 480$^2$ & 937.0 & 840.5 & - & 85.4 \\
 \hline
 ViT-B/16~\cite{dosovitskiy2020image} & 384$^2$ & 86.0 & 55.4 & 99 & 84.0 \\
 ViT-L/16~\cite{dosovitskiy2020image} & 384$^2$ & 307.0 & 190.7 & 30 & 85.2 \\
 \hline
 Swin-Base~\cite{liu2021swin} & 224$^2$/224$^2$  & 88.0 & 15.4 & 291 & 85.2 \\    
 \rowcolor{Gray}
 FocalNet-B  & 224$^2$/224$^2$ & 88.1 & 15.3 & 280 & \textbf{85.6}  \\
 Swin-Base~\cite{liu2021swin} & 384$^2$/384$^2$ & 88.0 & 47.1 & 91 & 86.4 \\    
 \rowcolor{Gray}
 FocalNet-B & 224$^2$/384$^2$ & 88.1 & 44.8 & 94 & \textbf{86.5}  \\
 Swin-Large~\cite{liu2021swin} & 224$^2$/224$^2$ & 196.5 & 34.5 & 155 & 86.3 \\    
 \rowcolor{Gray}
 FocalNet-L  & 224$^2$/224$^2$ & 197.1 & 34.2 & 144 & \textbf{86.5}  \\
 Swin-Large~\cite{liu2021swin} & 384$^2$/384$^2$ & 196.5 & 104.0 & 49 & \textbf{87.3} \\ 
 \rowcolor{Gray}
 FocalNet-L  & 224$^2$/384$^2$ & 197.1 & 100.6 & 50 & \textbf{87.3}  \\   
     \bottomrule
    \end{tabular}
    }
    \vspace{1mm}
    \captionsetup{font=footnotesize}
    \caption{ImageNet-1K finetuning results with models pretrained on ImageNet-22K. Numbers before and after ``/'' are resolutions used for pretraining and finetuning, respectively.}
    \label{tab:in22k}
\end{minipage}
\vspace{-3mm}
\end{minipage}
\end{table}

\textbf{ImageNet-22K pretraining.} We investigate the effectiveness of FocalNets when pretrained on ImageNet-22K which contains 14.2M images and 21K categories. Training details are described in the appendix. We report the results in Table~\ref{tab:in22k}. Though FocalNet-B/L are both pretrained with $224\times 224$ resolution and directly transferred to target domain with $384\times384$ image size, we can see that they consistently outperform Swin Transformers.

\subsection{Language-Image Contrast Learning}
We also study FocalNet in the recently popular language-image contrastive learning paradigm. More specifically, our experiment setting follows the Academic Track of the Image Classification in the Wild (ICinW) Challenge, where ImageNet-21K (ImageNet-1K images are removed), GCC3M+12M and YFCC15M are used in pre-training, and 20 downstream datasets and ImageNet-1K are evaluated to report the zero-shot performance~\cite{li2022elevater}. This setting was originally proposed in UniCL~\cite{yang2022unicl}.  We pre-train the model with the UniCL objective, and the vision backbone is specified as FocalNet-B and Swin-B for comparisons. The results are reported in Table~\ref{tab:elevater_icinw}. FocalNet-B outperforms Swin-B by 0.8 averaged gain on 20 datasets in ICinW and 2.0 gain on ImageNet-1K, respectively. 

\begin{table}[t]
\linespread{1}
\aboverulesep = 0.2em \belowrulesep = 0.2em
\scriptsize
\centering
\vspace{-3mm}
\resizebox{\textwidth}{!}{%
\begin{tabular}{cc|cccccccccccccccccccc|c|c} 
\bottomrule[1.5pt]
        \rotatebox[origin=lb]{90}{\smash{Dataset}}& \hspace{-1.em} &
        
        \rotatebox[origin=lb]{90}{\smash{Caltech101}} & \rotatebox[origin=lb]{90}{\smash{CIFAR10}} & \rotatebox[origin=lb]{90}{\smash{CIFAR100}} & \rotatebox[origin=lb]{90}{\smash{Country211}}  & \rotatebox[origin=lb]{90}{\smash{DescriTextures}}  & \rotatebox[origin=lb]{90}{\smash{EuroSAT}} & \rotatebox[origin=lb]{90}{\smash{FER2013}} & \rotatebox[origin=lb]{90}{\smash{FGVC Aircraft}} & \rotatebox[origin=lb]{90}{\smash{Food101}} & \rotatebox[origin=lb]{90}{\smash{GTSRB}} & \rotatebox[origin=lb]{90}{\smash{HatefulMemes}} & \rotatebox[origin=lb]{90}{\smash{KITTI}} & \rotatebox[origin=lb]{90}{\smash{MNIST}} & \rotatebox[origin=lb]{90}{\smash{Oxford Flowers}} & \rotatebox[origin=lb]{90}{\smash{Oxford Pets}}  & \rotatebox[origin=lb]{90}{\smash{PatchCamelyon}} & \rotatebox[origin=lb]{90}{\smash{Rendered SST2}} & \rotatebox[origin=lb]{90}{\smash{RESISC45}} & \rotatebox[origin=lb]{90}{\smash{Stanford Cars}} & \rotatebox[origin=lb]{90}{\smash{VOC2007}} & \rotatebox[origin=lb]{90}{\smash{\bf Mean Acc.~}} & 
        \rotatebox[origin=lb]{90}{\smash{ImageNet-1K}} \\
        \midrule
        
        Swin-B & \hspace{-1.2em} & 
        
  \hspace{-0.9em} 84.0 \hspace{-0.4em} &  \hspace{-0.9em} 93.0 \hspace{-0.4em} &  \hspace{-0.9em} 69.5 \hspace{-0.4em} &  \hspace{-0.9em} 7.3 \hspace{-0.4em} &  \hspace{-0.9em} 25.5 \hspace{-0.4em} &  \hspace{-0.9em} 24.4 \hspace{-0.4em} &  \hspace{-0.9em} 30.4 \hspace{-0.4em} &  \hspace{-0.9em} 2.7 \hspace{-0.4em} &  \hspace{-0.9em} 71.0 \hspace{-0.4em} &  \hspace{-0.9em} 9.0 \hspace{-0.4em} &  \hspace{-0.9em} 52.6 \hspace{-0.4em} &  \hspace{-0.9em} 12.4 \hspace{-0.4em} &  \hspace{-0.9em} 10.1 \hspace{-0.4em} &  \hspace{-0.9em} 70.4 \hspace{-0.4em} &  \hspace{-0.9em} 52.4 \hspace{-0.4em} &  \hspace{-0.9em} 50.6 \hspace{-0.4em} &  \hspace{-0.9em} 50.1 \hspace{-0.4em} &  \hspace{-0.9em} 44.8 \hspace{-0.4em} &  \hspace{-0.9em} 13.8 \hspace{-0.4em} &  \hspace{-0.9em} 81.3 \hspace{-0.4em} &   
          \hspace{-0.9em}43.2\hspace{-0.4em} &
           \hspace{-0.9em}52.2 \\
         
         FocalNet-B & \hspace{-1.2em} & 
        
         \hspace{-0.9em} 84.8 \hspace{-0.4em} &  \hspace{-0.9em} 90.2 \hspace{-0.4em} &  \hspace{-0.9em} 67.8 \hspace{-0.4em} &  \hspace{-0.9em} 6.7 \hspace{-0.4em} &  \hspace{-0.9em} 25.4 \hspace{-0.4em} &  \hspace{-0.9em} 35.3 \hspace{-0.4em} &  \hspace{-0.9em} 30.8 \hspace{-0.4em} &  \hspace{-0.9em} 3.5 \hspace{-0.4em} &  \hspace{-0.9em} 68.3 \hspace{-0.4em} &  \hspace{-0.9em} 11.1 \hspace{-0.4em} &  \hspace{-0.9em} 51.0 \hspace{-0.4em} &  \hspace{-0.9em} 17.9 \hspace{-0.4em} &  \hspace{-0.9em} 11.3 \hspace{-0.4em} &  \hspace{-0.9em} 71.7 \hspace{-0.4em} &  \hspace{-0.9em} 44.9 \hspace{-0.4em} &  \hspace{-0.9em} 52.1 \hspace{-0.4em} &  \hspace{-0.9em} 49.5 \hspace{-0.4em} &  \hspace{-0.9em} 41.4 \hspace{-0.4em} &  \hspace{-0.9em} 24.2 \hspace{-0.4em} &  \hspace{-0.9em} 81.3 \hspace{-0.4em} &  
          \hspace{-0.9em}{\bf44.0}\hspace{-0.4em} &
          \hspace{-0.9em}54.2 \\
          
        Gains&  \hspace{-1.2em} &  
        \hspace{-0.9em}\textcolor{green!50!black}{\bf0.9}\hspace{-0.4em} &  \hspace{-0.9em} -2.7 \hspace{-0.4em} &  \hspace{-0.9em} -1.7 \hspace{-0.4em} &  \hspace{-0.9em} -0.6 \hspace{-0.4em} &  \hspace{-0.9em} -0.1 \hspace{-0.4em} &  \hspace{-0.9em}\textcolor{green!50!black}{\bf11.0}\hspace{-0.4em} &  \hspace{-0.9em}\textcolor{green!50!black}{\bf0.5}\hspace{-0.4em} &  \hspace{-0.9em}\textcolor{green!50!black}{\bf0.8}\hspace{-0.4em} &  \hspace{-0.9em} -2.7 \hspace{-0.4em} &  \hspace{-0.9em}\textcolor{green!50!black}{\bf2.1}\hspace{-0.4em} &  \hspace{-0.9em} -1.6 \hspace{-0.4em} &  \hspace{-0.9em}\textcolor{green!50!black}{\bf5.5}\hspace{-0.4em} &  \hspace{-0.9em}\textcolor{green!50!black}{\bf1.2}\hspace{-0.4em} &  \hspace{-0.9em}\textcolor{green!50!black}{\bf1.3}\hspace{-0.4em} &  \hspace{-0.9em} -7.6 \hspace{-0.4em} &  \hspace{-0.9em}\textcolor{green!50!black}{\bf1.6}\hspace{-0.4em} &  \hspace{-0.9em} -0.6 \hspace{-0.4em} &  \hspace{-0.9em} -3.4 \hspace{-0.4em} &  \hspace{-0.9em}\textcolor{green!50!black}{\bf10.5}\hspace{-0.4em} &  \hspace{-0.9em}\textcolor{green!50!black}{\bf0.0}\hspace{-0.4em} &  
        \hspace{-0.9em}\textcolor{green!50!black}{\bf+0.8}\hspace{-0.4em} &  
        \hspace{-0.9em}\textcolor{green!50!black}{\bf2.0}\hspace{-0.4em}   
        \\    
\bottomrule[1.0pt]	
\end{tabular}%
}
\caption{Zero-shot performance comparison of FocalNet-B and Swin-B on ELEVATER benchmark~\cite{li2022elevater}. We calculate the gains marked in green for positive results. The mean score over 20 datasets and the top-1 accuracy on ImageNet-1K are reported in the last two columns, respectively.}
\label{tab:elevater_icinw}
\vspace{-0mm}
\end{table}

\subsection{Detection and Segmentation}

\textbf{Object detection and instance segmentation}. We make comparisons on object detection with COCO~2017~\cite{lin2014microsoft}. We choose Mask R-CNN~\cite{he2017mask} as the detection method and use FocalNet-T/S/B pretrained on ImageNet-1K as the backbones. All models are trained on the 118k training images and evaluated on 5K validation images. 
We use two standard training recipes, $1\times$ schedule with 12 epochs and $3\times$ schedule with 36 epochs. Following~\cite{liu2021swin}, we use the same multi-scale training strategy by randomly resizing the shorter side of an image to $[480,800]$. Similar to~\cite{yang2021focal}, we increase the kernel size $k^{\ell}$ by 6 for context aggregation at all focal levels to adapt to higher input resolutions. Instead of up-sampling the relative position biases as in~\cite{yang2021focal}, FocalNets uses simple zero-padding for the extra kernel parameters. This expanding introduces negligible overhead but helps extract longer range contexts. For training, we use AdamW~\cite{loshchilov2017decoupled} as the optimizer with initial learning rate $10^{-4}$ and weight decay $0.05$. All models are trained with batch size 16. We set the stochastic drop rates to $0.1$, $0.2$, $0.3$ in $1\times$ and $0.3$, $0.5$, $0.5$ in $3\times$ training schedule for FocalNet-T/S/B, respectively. 

The results are shown in Table~\ref{tab:maskrcnn}. We measure both box and mask mAP, and report the results for both small and large receptive field models. Comparing with Swin Transformer, FocalNets improve the box mAP (AP$^b$) by $2.2$, $1.5$ and $1.9$ in $1\times$ schedule for tiny, small and base models, respectively. In $3\times$ schedule, the improvements are still consistent and significant. Remarkably, the $1\times$ performance of FocalNet-T/B (45.9/48.8) rivals Swin-T/B (46.0/48.5) trained with $3\times$ schedule. When comparing with FocalAtt~\cite{yang2021focal}, FocalNets with large receptive fields consistently outperform under all settings and cost much less FLOPs. For instance segmentation, we observe the similar trend as that of object detection for FocalNets.
To further verify the generality of FocalNets, we train three detection models, Cascade Mask R-CNN~\cite{cai2018cascade}, Sparse RCNN~\cite{sun2020sparse} and  ATSS~\cite{zhang2020bridging} with FocalNet-T as the backbone. We train all models with $3\times$ schedule, and report the box mAPs in Table~\ref{tab:ablation_on_detectors}. As we can see, FocalNets bring clear gains to all three detection methods over the previous SoTA methods.

\begin{table}[t]
\centering
\resizebox{0.99\linewidth}{!}{
\setlength{\tabcolsep}{1.9pt}
\footnotesize
\begin{tabular}{lcc|lccccc|lccccc}
\toprule
\multirow{2}{*}{Backbone} & \#Params & FLOPs & \multicolumn{6}{c}{Mask R-CNN 1x} & \multicolumn{6}{c}{Mask R-CNN 3x}\\
\cmidrule{4-15}
 & (M) & (G) & $AP^b$ & $AP^b_{50}$ & $AP^b_{75}$ & $AP^m$ & $AP^m_{50}$ & $AP^m_{75}$ & $AP^b$ & $AP^b_{50}$ & $AP^b_{75}$ & $AP^m$ & $AP^m_{50}$ & $AP^m_{75}$\\
\midrule
ResNet50~\cite{he2016deep} & 44.2 & 260 & 38.0 & 58.6 & 41.4 & 34.4 & 55.1 & 36.7 & 41.0 & 61.7 & 44.9 & 37.1 & 58.4 & 40.1 \\
PVT-Small\cite{wang2021pyramid} & 44.1 & 245 & 40.4 & 62.9 & 43.8 & 37.8 & 60.1 & 40.3 & 43.0 & 65.3 & 46.9 & 39.9 & 62.5 & 42.8 \\
Twins-SVT-S~\cite{chu2021twins} & 44.0 & 228 & 43.4 & 66.0 & 47.3 & 40.3 & 63.2 & 43.4 & 46.8 & 69.2 & 51.2 & 42.6 & 66.3 & 45.8 \\
Swin-Tiny~\cite{liu2021swin} & 47.8 & 264 & 43.7 & 66.6 & 47.7 & 39.8 & 63.3 & 42.7 & 46.0 & 68.1 & 50.3 & 41.6 & 65.1 & 44.9 \\
\rowcolor{Gray}
FocalNet-T~(SRF) & 48.6 & 267  & \textbf{45.9}{\textcolor{green!50!black}{(\bf+2.2)}} & \textbf{68.3} & \textbf{50.1} & \textbf{41.3} & \textbf{65.0} & \textbf{44.3} & \textbf{47.6}{\textcolor{green!50!black}{(\bf+1.6)}} & \textbf{69.5} & \textbf{52.0} & \textbf{42.6} & \textbf{66.5} & \textbf{45.6} \\
FocalAtt-Tiny~\cite{yang2021focal} & 48.8 & 291 & {44.8} & {67.7} & {49.2} & {41.0} & {64.7} & {44.2} & {47.2} & {69.4} & {51.9} & {42.7} & {66.5} & {45.9} \\
\rowcolor{Gray}
FocalNet-T~(LRF) & 48.9 & 268 & \textbf{46.1}{\textcolor{green!50!black}{(\bf+1.3)}} & \textbf{68.2} & \textbf{50.6} & \textbf{41.5} & \textbf{65.1} & \textbf{44.5} & \textbf{48.0}{\textcolor{green!50!black}{(\bf+0.8)}} & \textbf{69.7} & \textbf{53.0} & \textbf{42.9} & \textbf{66.5} & \textbf{46.1} \\
\midrule
ResNet101~\cite{he2016deep} & 63.2 & 336 & 40.4 & 61.1 & 44.2 & 36.4 & 57.7 & 38.8 & 42.8 & 63.2 & 47.1 & 38.5 & 60.1 & 41.3 \\
ResNeXt101-32x4d~\cite{xie2017aggregated} & 62.8 & 340& 41.9 & 62.5 & 45.9 & 37.5 & 59.4 & 40.2 & 44.0 & 64.4 & 48.0 & 39.2 & 61.4 & 41.9 \\
PVT-Medium~\cite{wang2021pyramid} & 63.9 & 302 & 42.0 & 64.4 & 45.6 & 39.0 & 61.6 & 42.1 & 44.2 & 66.0 & 48.2 & 40.5 & 63.1 & 43.5 \\
Twins-SVT-B~\cite{chu2021twins} & 76.3 & 340 & 45.2 & 67.6 & 49.3 & 41.5 & 64.5 & 44.8 & 48.0 & 69.5 & 52.7 & 43.0 & 66.8 & 46.6 \\
Swin-Small~\cite{liu2021swin} & 69.1 & 354 & 46.5 & 68.7 & 51.3 & 42.1 & 65.8 & 45.2 & 48.5 & 70.2 & 53.5 & 43.3 & 67.3 & 46.6 \\
\rowcolor{Gray}
FocalNet-S~(SRF) & 70.8 & 356 & \textbf{48.0}{\textcolor{green!50!black}{(\bf+1.5)}} & \textbf{69.9} & \textbf{52.7} & \textbf{42.7} & \textbf{66.7} & \textbf{45.7} &  \textbf{48.9}{\textcolor{green!50!black}{(\bf+0.4)}} & \textbf{70.1} & \textbf{53.7} & \textbf{43.6} & 67.1 & \textbf{47.1} \\
FocalAtt-Small~\cite{yang2021focal} & 71.2 & 401 &{47.4} & {69.8} & {51.9} & {42.8} & {66.6} & {46.1} & {48.8} & {70.5} & {53.6} & \textbf{43.8} & {67.7} & {47.2} \\
\rowcolor{Gray}
FocalNet-S~(LRF) & 72.3 & 365 & \textbf{48.3}{\textcolor{green!50!black}{(\bf+0.9)}} & \textbf{70.5} & \textbf{53.1} & \textbf{43.1} & \textbf{67.4} & \textbf{46.2} & \textbf{49.3}{\textcolor{green!50!black}{(\bf+0.5)}} & \textbf{70.7} & \textbf{54.2} & \textbf{43.8} & \textbf{67.9} & \textbf{47.4} \\
\midrule
ResNeXt101-64x4d~\cite{xie2017aggregated} & 102.0 & 493 & 42.8 & 63.8 & 47.3 & 38.4 & 60.6 & 41.3 & 44.4 & 64.9 & 48.8 & 39.7 & 61.9 & 42.6 \\
PVT-Large\cite{wang2021pyramid} & 81.0 & 364 & 42.9 & 65.0 & 46.6 & 39.5 & 61.9 & 42.5 & 44.5 & 66.0 & 48.3 & 40.7 & 63.4 & 43.7 \\
Twins-SVT-L~\cite{chu2021twins} & 119.7 & 474 & 45.9 & - & - & 41.6 & - & - & - & - & - & - & - & -   \\
Swin-Base~\cite{liu2021swin} & 107.1 & 497 & 46.9 & 69.2 & 51.6 & 42.3 & 66.0 & 45.5 & 48.5 & 69.8 & 53.2 & 43.4 & 66.8 & 46.9 \\
\rowcolor{Gray}
FocalNet-B~(SRF) & 109.4 & 496 &  \textbf{48.8}{\textcolor{green!50!black}{(\bf+1.9)}} & \textbf{70.7} & \textbf{53.5} & \textbf{43.3} & \textbf{67.5} & \textbf{46.5} & \textbf{49.6}{\textcolor{green!50!black}{(\bf+1.1)}} & \textbf{70.6} & \textbf{54.1} & \textbf{44.1} & \textbf{68.0} & \textbf{47.2} \\
FocalAtt-Base~\cite{yang2021focal} & 110.0 & 533 & {47.8} & {70.2} & {52.5} & {43.2} & {67.3} & {46.5} & {49.0} & {70.1} & {53.6} & {43.7} & {67.6} & {47.0} \\
\rowcolor{Gray}
FocalNet-B~(LRF) & 111.4 & 507  & \textbf{49.0}{\textcolor{green!50!black}{(\bf+1.2)}} & \textbf{70.9} & \textbf{53.9} & \textbf{43.5} & \textbf{67.9} & \textbf{46.7} & \textbf{49.8}{\textcolor{green!50!black}{(\bf+0.8)}} & \textbf{70.9} & \textbf{54.6} & \textbf{44.1} & \textbf{68.2} & \textbf{47.2}  \\
\bottomrule
\end{tabular}
}
\captionsetup{font=footnotesize}    
\caption{COCO object detection and instance segmentation results with Mask R-CNN~\cite{he2017mask}. 
}
\label{tab:maskrcnn}
\vspace{-6mm}
\end{table}

\begin{table}[t]
\begin{minipage}{0.56\linewidth}
\resizebox{1.0\linewidth}{!}{
\centering
\footnotesize
\setlength{\tabcolsep}{2.5pt}
  \begin{tabular}{cl|ccclll}
    \toprule
    Method & Backbone & \#Param. & FLOPs & $AP^b$ & $AP^b_{50}$ & $AP^b_{75}$ \\
    \midrule	 
    \multirow{5}{*}{{C. Mask R-CNN~\cite{cai2018cascade}}} 
         & R-50~\cite{he2016deep} & 82.0 & 739 & 46.3 & 64.3 & 50.5 \\
         & DW-Net-T~\cite{han2021demystifying} & 82.0 & 730 & 49.9 & 68.6 & 54.3 \\
         & Swin-T~\cite{liu2021swin} & 85.6 & 742 & 50.5 & 69.3 & 54.9 \\
         & \cellcolor{Gray}FocalNet-T~(SRF) 
         & \cellcolor{Gray}86.4 
         & \cellcolor{Gray}746 
         & \cellcolor{Gray}\textbf{51.5} 
         & \cellcolor{Gray}\textbf{70.1} 
         & \cellcolor{Gray}\textbf{55.8} \\
         & FocalAtt-T~\cite{yang2021focal} & 86.7 & 770 & {51.5} & \textbf{70.6} & {55.9} \\
           \rowcolor{Gray}    \cellcolor{white}
         & FocalNet-T~(LRF) & 87.1 & 751 & \textbf{51.5} & {70.3} & \textbf{56.0} \\
    \midrule
    \multirow{5}{*}{Sparse R-CNN~\cite{sun2020sparse}} 
         & R-50~\cite{he2016deep} & 106.1 & 166 & 44.5 & 63.4 & 48.2 \\
         & Swin-T~\cite{liu2021swin} & 109.7 & 172 & 47.9 & 67.3 & 52.3 \\
         & \cellcolor{Gray}FocalNet-T~(SRF) 
         & \cellcolor{Gray}110.5 
         & \cellcolor{Gray}172 
         & \cellcolor{Gray}\textbf{49.6} 
         & \cellcolor{Gray}\textbf{69.1} 
         & \cellcolor{Gray}\textbf{54.2} \\            
         & FocalAtt-T~\cite{yang2021focal} & 110.8 & 196 & {49.0} & {69.1} & {53.2} \\
    \rowcolor{Gray}    \cellcolor{white}   
         & FocalNet-T~(LRF) & 111.2 & 178 & \textbf{49.9} & \textbf{69.6} & \textbf{54.4}   \\
    \midrule
     \multirow{5}{*}{ATSS~\cite{zhang2020bridging}} 
         & R-50~\cite{he2016deep} & 32.1 & 205 & 43.5 & 61.9 & 47.0 \\
         & Swin-T~\cite{liu2021swin} & 35.7 & 212 & 47.2 & 66.5 & 51.3 \\
         & \cellcolor{Gray}FocalNet-T~(SRF) 
         & \cellcolor{Gray}36.5 
         & \cellcolor{Gray}215 
         & \cellcolor{Gray}\textbf{49.2} 
         & \cellcolor{Gray}\textbf{68.1} 
         & \cellcolor{Gray}\textbf{54.2} \\           
         & FocalAtt-T~\cite{yang2021focal} & 36.8 & 239 & {49.5} & \textbf{68.8} & {53.9}\\
           \rowcolor{Gray}    \cellcolor{white}  
         & FocalNet-T~(LRF) & 37.2 & 220 & \textbf{49.6} & 68.7 & \textbf{54.5} \\         
 \bottomrule
  \end{tabular}}
    \captionsetup{font=footnotesize}    
    \caption{A comparison of models with different object detection methods, trained using the $3\times$ schedule.}
    \label{tab:ablation_on_detectors}
    \vspace{-4mm}
\end{minipage}
\quad
\begin{minipage}{0.4\linewidth}
\setlength{\tabcolsep}{1.8pt}
\centering
\footnotesize
\resizebox{0.99\linewidth}{!}{
  \begin{tabular}{l|ccccc}
    \toprule
    Backbone & Crop Size & \#Param. & FLOPs & mIoU & +MS \\
    \midrule	 
    ResNet-101~\cite{he2016deep} & 512 & 86 & 1029 & 44.9 & - \\
    Twins-SVT-L~\cite{chu2021twins} & 512 & 133 & - & 48.8 & 50.2 \\
    DW-Net-T~\cite{han2021demystifying} & 512 & 56 & 928 & 45.5 & - \\
    DW-Net-B~\cite{han2021demystifying} & 512 & 132 & 924 & 48.3 & - \\
    \midrule
    Swin-T~\cite{liu2021swin} & 512 & 60 & 941 & 44.5 & 45.8  \\
    \rowcolor{Gray}    
    FocalNet-T~(SRF) & 512 & 61 & 944 & \textbf{{46.5}} & \textbf{47.2} \\    
    FocalAtt-T~\cite{yang2021focal} & 512 & 62 & 998 & 45.8 & 47.0 \\
    \rowcolor{Gray}    
    FocalNet-T~(LRF) & 512 & 61 & 949 & \textbf{46.8} & \textbf{47.8} \\    
    Swin-S~\cite{liu2021swin} & 512 & 81 & 1038 & 47.6 & 49.5  \\
    \rowcolor{Gray}   
    FocalNet-S~(SRF) & 512 & 83 & 1035 & \textbf{49.3} & \textbf{50.1} \\     
    FocalAtt-S~\cite{yang2021focal} & 512 & 85 & 1130 & 48.0 & \textbf{50.0}\\     
    \rowcolor{Gray}   
    FocalNet-S~(LRF) & 512 & 84 & 1044 & \textbf{{49.1}} & \textbf{50.1} \\     
    Swin-B~\cite{liu2021swin} & 512 & 121 & 1188 & 48.1 & 49.7  \\    
    \rowcolor{Gray}    
    FocalNet-B~(SRF) & 512 & 124 & 1180 & \textbf{50.2} & \textbf{51.1} \\    
    FocalAtt-B~\cite{yang2021focal} & 512 & 126 & 1354 & 49.0 & 50.5 \\  
    \rowcolor{Gray}    
    FocalNet-B~(LRF) & 512 & 126 & 1192 & \textbf{50.5} & \textbf{51.4} \\    
    \bottomrule
  \end{tabular} 
  }
  \captionsetup{font=footnotesize}    
  \caption{Semantic segmentation on ADE20K~\cite{zhou2017scene}. All models are trained with UperNet~\cite{xiao2018unified}. MS means multi-scale evaluation.}
  \label{tab:semantic_segmentation}
  \end{minipage} 
\vspace{-5mm}
\end{table}

\textbf{Semantic Segmentation}. We benchmark FocalNets on semantic segmentation, a dense prediction task that requires fine-grained understanding and long-range interactions. We use ADE20K~\cite{zhou2017scene} for our experiments and follow~\cite{liu2021swin} to use UperNet~\cite{xiao2018unified} as the segmentation method. With FocalNet-T/S/B trained on ImageNet-1K as the backbones, we train UperNet for 160k iterations with input resolution $512\times 512$ and batch size 16. For comparisons, we report both single- and multi-scale (MS) mIoU. Table~\ref{tab:semantic_segmentation} shows the results with different backbones. FocalNet outperforms Swin and Focal Transformer significantly under all settings. Even for the base models,  FocalNet (SRF) exceeds Swin Transformer by $2.1$ and $1.4$ at single- and multi-scale, respectively. 
Compared with Focal Transformer, FocalNets outperform Focal Transformer, with a larger gain than that of Swin Transformer, and consume much less FLOPs. These results demonstrate the superiority of FocalNets on the pixel-level dense prediction tasks, in addition to the instance-level object detection task.

\begin{table}[t]
\begin{minipage}{0.47\linewidth}
\centering
\footnotesize
\resizebox{1.0\linewidth}{!}{
  \begin{tabu}{lcc|cc}
    \toprule
    Backbone & Method & \#Param  & mIoU & +MS \\
    \midrule	 
    HRNet-w48~\cite{SunXLW19} & OCRNet~\cite{yuan2019object} & 71M  & 45.7 & -\\
    ResNeSt-200~\cite{zhang2020resnest} & DLab.v3+~\cite{chen2018encoder} & 88M & 48.4 & -\\
    \midrule
    Swin-B~\cite{liu2021swin} & UperNet~\cite{xiao2018unified} & 121M & 48.1 & 49.7  \\
    Twins-SVT-L~\cite{chu2021twins} & UperNet~\cite{xiao2018unified} & 133M & 48.8 & 50.2 \\
    MiT-B5~\cite{xie2021segformer} & SegFormer~\cite{xie2021segformer} & 85M & 51.0 & 51.8\\
    ViT-L/16$^{\dagger}$~\cite{dosovitskiy2020image} & SETR~\cite{zheng2020rethinking} & 308M & 50.3 & - \\    
    Swin-L$^{\dagger}$~\cite{liu2021swin} & UperNet~\cite{xiao2018unified} & 234M & 52.1 & 53.5 \\
    ViT-L/16$^{\dagger}$~\cite{dosovitskiy2020image} & Segmenter~\cite{strudel2021segmenter} & 334M & 51.8 & 53.6 \\
    Swin-L$^{\dagger}$~\cite{liu2021swin} & K-Net~\cite{zhang2021knet} & - & - & 54.3 \\
    Swin-L$^{\dagger}$~\cite{liu2021swin} & PatchDiverse~\cite{gong2021vision} & 234M & 53.1 & 54.4 \\
    VOLO-D5~\cite{yuan2021volo} & UperNet~\cite{xiao2018unified} & - & - & 54.3 \\
    Focal-L$^{\dagger}$ & UperNet~\cite{xiao2018unified} & 240M & 54.0 & 55.4 \\   
    CSwin-L$^{\dagger}$ & UperNet~\cite{xiao2018unified} & 208M & 54.0 & 55.7 \\
    \midrule
    \rowfont{\color{gray!60}}
    {BEIT-L$^{\dagger}$} & UperNet~\cite{xiao2018unified} & 441M & 56.7 & 57.0 \\
    \rowfont{\color{gray!60}}
    Swinv2-G$^{\ddagger}$~\cite{liu2021swinv2} &
    UperNet~\cite{xiao2018unified} & >3.0B & 59.1 & - \\
    \rowfont{\color{gray!60}}
    ViT-Adapter-L$^{\dagger}$~\cite{chen2022vision} & Mask2Former~\cite{cheng2021masked} & 568M & 58.3 & 59.0 \\
    \midrule
    Swin-L$^{\dagger}$ & Mask2Former~\cite{cheng2021masked} & 216M & 56.4 & 57.7 \\
    Swin-L-FaPN$^{\dagger}$ & Mask2Former~\cite{cheng2021masked} & - & 56.1 & 57.3 \\
    Swin-L-SeMask$^{\dagger}$~\cite{jain2021semask}  & Mask2Former~\cite{cheng2021masked} & - & 57.0 & 58.2 \\
    \rowcolor{Gray}    
    FocalNet-L$^{\dagger}$ (Ours) & Mask2Former~\cite{cheng2021masked} & 218M & \textbf{57.3} & \textbf{58.5} \\
    \bottomrule
  \end{tabu} 
}
\captionsetup{font=footnotesize}    
\caption{Systematic comparisons of semantic segmentation on ADE20K validation set. $^{\dagger}$ indicates pretraining with ImageNet-22K and $^{\ddagger}$ means using extra data additionally. 
}
\vspace{-6mm}
\label{tab:ade20k_bigmodels}
\end{minipage}
\quad
\begin{minipage}{0.52\linewidth}
\centering
\footnotesize
\resizebox{0.98\linewidth}{!}{
  \begin{tabular}{lcc|ccc}
    \toprule
    Backbone  & Method & \#Param. & PQ & AP & mIoU \\
    \midrule	 
    ResNet-50~\cite{he2016deep} & DETR~\cite{carion2020end} & - & 43.4 & - & - \\
    ResNet-50~\cite{he2016deep} & K-Net~\cite{zhang2021knet} & - & 47.1 & - & - \\
    ResNet-50~\cite{he2016deep} & \makecell{Panoptic \\ SegFormer~\cite{li2021panoptic}} & 47M & 50.0 & - & - \\
    ResNet-50~\cite{he2016deep} & Mask2Former~\cite{cheng2021masked} & 44M & 51.9 & 41.7 & 62.4 \\
    \midrule
    PVTv2-B5~\cite{wang2022pvtv2} & \makecell{Panoptic \\ SegFormer~\cite{li2021panoptic}} & 101M & 54.1 & - & - \\
    Swin-T~\cite{liu2021swin} & MaskFormer~\cite{cheng2021per} & 42M & 47.7 & 33.6 & 60.4  \\
    Swin-B~\cite{liu2021swin} & MaskFormer~\cite{cheng2021per} & 102M & 51.1 & 37.8 & 62.6  \\    
    Swin-T~\cite{liu2021swin} & Mask2Former~\cite{cheng2021masked} & 47M & 53.2 & 43.3 & 63.2  \\
    Swin-B~\cite{liu2021swin} & Mask2Former~\cite{cheng2021masked} & 107M & 55.1 & 45.2 & 65.1  \\    
    \midrule
    Swin-L$\dagger$~\cite{liu2021swin} & MaskFormer~\cite{cheng2021per} & 212M & 52.7 & 40.1 & 64.8  \\
    {Swin-L$\dagger$~\cite{liu2021swin}} & \makecell{Panoptic \\ SegFormer~\cite{li2021panoptic}} & - & 55.8 & - & -  \\  
    Swin-L$\dagger$~\cite{liu2021swin} & \makecell{Mask2Former~\cite{cheng2021per} \\ (200 queries)} & 216M & 57.8 & \textbf{48.6} & \textbf{67.4}  \\     
    FocalNet-L$\dagger$~(Ours) & \makecell{Mask2Former~\cite{cheng2021per} \\ (200 queries)} & 226M & \textbf{57.9} & 48.4 & 67.3  \\ 
    \bottomrule
  \end{tabular} 
  }
  \captionsetup{font=footnotesize}    
  \caption{Panoptic segmentation on COCO~\cite{lin2014microsoft}. $\dagger$ means pretraining with ImageNet-22K. All models evaluated on minival with single-scale. PQ, AP and mIoU are three metrics for measuring the panoptic segmentation, instance segmentation and semantic segmentation, respectively.}
  \label{tab:panoptic_segmentation}    
\end{minipage}
\end{table}
\begin{table}[t]
\centering
\resizebox{1.0\linewidth}{!}{
    \begin{tabular}{lcccc|cccc}
    \toprule
         \multirow{2}{*}{Method}  &  \multirow{2}{*}{\#Params.} & \multirow{2}{*}{Backbone Pretraining} & \multirow{2}{*}{Detection Pretraining} & \multirow{2}{*}{W/ Mask} & \multicolumn{2}{c}{val2017} & \multicolumn{2}{c}{test-dev} \\
         & & & & & w/o TTA & w/ TTA & w/o TTA & w/ TTA \\
    \midrule
         Swin-L (HTC++)~\cite{liu2021swin}  & 284M & IN-22K~(14M) & n/a & \checkmark & 57.1 & 58.0 & 57.7 & 58.7 \\
         DyHead (Swin-L)~\cite{dai2021dynamic}  & 213M & IN-22K~(14M) & n/a & \cmark & 56.2 & 58.4 & - & - \\
         Focal-L (DyHead)~\cite{yang2021focal}  & 229M & IN-22K~(14M) & n/a & \cmark & 56.4 & 58.7 & - & 58.9 \\
         Soft-Teacher (Swin-L)~\cite{xu2021end} & 284M & IN-22K (14M) & COCO-unlabeled + O365 &  \cmark & 60.1 & 60.7 & - & 61.3 \\
         GLIP (DyHead)~\cite{li2022grounded} & $\ge$284M & IN-22K (14M) & FourODs + GoldG + Cap24M & \xmark & - & 60.8 & - & 61.5 \\
         Florence-CoSwin-H~\cite{yuan2021florence} & $\ge$637M & FLD-900M (900M) & FLD-9M & \xmark & - & 62.0 & - & 62.4 \\
         SwinV2-G~\cite{liu2021swinv2} & 3.0B & In-22K + ext-70M (84M) & O365 & \cmark & 61.9 & 62.5 & - & 63.1 \\
         DINO (Swin-L)~\cite{zhang2022dino} & 218M & IN-22K (14M) & O365 & \xmark & 63.1 & 63.2 & 63.2 & 63.3 \\
         BEIT-3~\cite{wang2022image} & 1.9B & IN-22K + Image-Text Pairs (35M) + Text (160GB) & O365 & \cmark & - & - & - & 63.7 \\
         FD-SwinV2-G~\cite{wei2022contrastive} & 3.0B & IN-22K + IN-1K + ext-70M (85M)  & O365 & \cmark & - & - & - & 64.2 \\
         \rowcolor{Gray}    
         FocalNet-H (DINO) & 746M & IN-22K (14M) & O365 & \xmark & \textbf{64.0} &\textbf{ 64.2} & \textbf{64.1} &\textbf{ 64.4} \\
         \bottomrule
    \end{tabular}
    }
    \caption{Comparisons of best detection models on COCO across the leaderboard. ``W/ Mask'' means whether using mask annotations for finetuning on COCO. ``TTA'' means test-time-augmentation. ``FLD-900M'' used in Florence~\cite{yuan2021florence} contains 900M free-form image-text pairs. ``ex-70M'' used in Swinv2~\cite{liu2021swinv2} is a private classification dataset containing 70M images.}
    \label{tab:object_detection_at_scale}
    \vspace{-7mm}
\end{table}

\paragraph{Scaling-up FocalNets.} Given the superior results for FocalNets on object detection and segmentation shown above, we further investigate its effectiveness while scaling up. Particularly, to fairly compare with Swin-L pretrained on ImageNet-22K with 384$\times$384, we also pretrain our FocalNet-L on ImageNet-22K with 384$\times$384 with 3 focal levels and kernel sizes $[3,5,7]$. We use Mask2former~\cite{cheng2021masked} for semantic segmentation on ADE20K and panoptic segmentation on COCO. As shown in Table~\ref{tab:ade20k_bigmodels}, FocalNet-L achieves superior performance to Swin-L with similar model size and same pretraining data. We note that the methods in gray font like Swinv2-G and ViT-Adapter-L achieve better performance but use much more parameters and training data. In Table~\ref{tab:panoptic_segmentation}, we compare different models for panoptic segmentation on COCO with 133 categories. {Our FocalNet-L slightly outperforms Swin-L on PQ}. These results clearly demonstrate the effectiveness of our FocalNets for various segmentation tasks when being scaled up to large model size.

Finally, we study the effectiveness of our FocalNets for object detection at scale. Currently, many methods scaled to billions of model parameters~\cite{liu2021swinv2, wang2022image}. To catch up, we pretrain a huge FocalNet with around 700M parameters on ImageNet-22K for 90 epochs using the same regime used for our large models. On top of the pretrained FocalNet-H, we use the public available object detection method DINO~\cite{zhang2022dino}. Following previous works, we pretrain the detection model on Object365~\cite{shao2019objects365}, and finetune it on COCO training set. We report the results in Table~\ref{tab:object_detection_at_scale}. Our model beats the SoTA methods like SwinV2-G~\cite{liu2021swinv2}, BEIT-3~\cite{zhang2022dino}, while consuming much less parameters and backbone pretraining data. 
It also outperforms the SwinV2-G tuned with the new feature distillation technique~\cite{wei2022contrastive}, and \textbf{establishes a new SoTA on the COCO leaderboard}. We believe our FocalNet can be further boosted after equipping the same technique and being pretrained on more data.

\subsection{Network Inspection}


\textbf{Model Variants}. 
We compare in Table~\ref{tab:model_variants} six different model variants derived from FocalNet.
\begin{itemize}[leftmargin=*]
    \item \textbf{Depth-wise ConvNet.} It feeds the feature vectors at the top level $L$ to a two-layer MLP. The resultant model is close to DW-Net \cite{han2021demystifying}. Although it can achieve 81.6\%, surpassing Swin (81.3\%), it underperforms FocalNet by 0.7\%. FocalNet uses depth-wise convolutions as a component but differently for aggregating contexts, which is then used to modulate each individual tokens.
    \item \textbf{Pooling Aggregator.} It replaces the depth-wise convolution module with average pooling, and is similar to MetaFormer~\cite{yu2021metaformer} in terms of token aggregation. Average pooling has slightly lower complexity but leads to a significant drop of accuracy by 1.8\%. Compared with depth-wise convolution, pooling is permutation-invariant and thus incapable of capturing visual structures.
    \item \textbf{Global Pooling Aggregator.} It removes local aggregations at all levels and only keeps the global one ($\Zmat^{L+1}$). This variant resembles SENet~\cite{hu2018squeeze}. It turns out that global context alone is insufficient for visual modeling, leading to a significant 6.7\% drop.
    \item \textbf{Multi-scale Self-Attention.} Given the summarized tokens at different levels, a straightforward way to combine them is performing a SA among all of them. We have developed two SA methods: computing $q,k,v$ before and after aggregation, respectively. Both methods result in visible performance drop and increase the run time latency, compared to FocalNet.
    \item \textbf{Sliding-window Self-Attention.}  Finally, we apply a sliding-window SA for each visual token within a window. Since it involves dense interactions for each fine-grained tokens, the time and memory cost explodes, and the performance is worse than FocalNet.
    \vspace{-2mm}
\end{itemize}

\begin{table}[t]
\centering
\setlength{\tabcolsep}{1.8pt}
\resizebox{0.98\linewidth}{!}{
    \footnotesize
    \begin{tabular}{ll|cccl}
    \toprule
    Model & Formula & {\#Param.} & {FLOPs} & {Throughput} & {Top-1} \\   
        \midrule
        
        \textbf{FocalNet-T~(LRF)} & $\yv_i = q(\xv_i) \odot h(\sum_{\ell=1}^{L+1} \gv^\ell_i \cdot \zv^\ell_i)$  & 28.6 & 4.49 & 696 & 82.3 \\
        
        \midrule
        
        $\rightarrow$ \textbf{Depth-width ConvNet} & $\yv_i = q(\GeLU(h(\zv^L_i))) $ & 28.6 & 4.47 & 738 & 81.6~\tiny{\textcolor{red}{(-0.7)}} \\
        
        \midrule
        
        $\rightarrow$ \textbf{Pooling Aggregator} & $\yv_i = q(\xv_i) \odot h(\sum_{\ell=1}^{L+1} \gv^\ell_i \cdot \AvgPool(\zv^{\ell-1}_i))$ & 28.3 & 4.37 & 676 & 80.5~\tiny{\textcolor{red}{(-1.8)}} \\  
        
        $\rightarrow$ \textbf{Global Pooling Aggregator} & $\yv_i = q(\xv_i) \odot h(\gv_i \cdot \AvgPool(f_z(X)))$ & 28.3 & 4.36 & 883 &  75.7~\tiny{\textcolor{red}{(-6.7)}} \\
        
        \midrule
        
        $\rightarrow$ \textbf{Multi-scale Self-Attention (QKV first)} & $\yv_i = MHSA(\xv_i, \zv^1_i,...,\zv^{L+1}_i), f_z, q,h=Identity(\cdot)$ & 28.6 & 4.61 & 456 & 81.5~\tiny{\textcolor{red}{(-0.8)}} \\
        
        $\rightarrow$ \textbf{Multi-scale Self-Attention (QKV later)} & $\yv_i = MHSA(\xv_i, \zv^1_i,...,\zv^{L+1}_i), f_z,q,h=Identity(\cdot)$ & 28.6 & 7.26 & 448 & 80.8~\tiny{\textcolor{red}{(-1.5)}} \\
        
        $\rightarrow$ \textbf{Sliding-window Self-Attention} & $\yv_i = MHSA(\xv_i, \mathcal{N}(\xv_i)), |\mathcal{N}(\xv_i)|=7\times 7-1 $ & 28.3 & 4.49 & 103 & 81.5~\tiny{\textcolor{red}{(-0.8)}} \\
        
        \bottomrule
    \end{tabular}
    }
    \captionsetup{font=footnotesize}    
    \caption{Performance for different FocalNet model variants.}
    \label{tab:model_variants}
    \vspace{-6mm}
\end{table}

\begin{table}[t]
\begin{minipage}{0.53\linewidth}
\setlength{\tabcolsep}{1.8pt}
\resizebox{0.99\linewidth}{!}{
    \footnotesize
    \centering
    \begin{tabular}{l|cclll}
    \toprule
       Model  & {FLOPs} & {Throughput} & {Top-1} & AP$^b$ & $AP^m$ \\
       \toprule
       FocalNet-T~(LRF) & 4.48 & 696 & 82.3 & 46.2 & 41.6 \\
       \midrule
        {Additive} & 4.49 & 670 &  81.5~\tiny{\textcolor{red}{(-0.8)}} & 45.6~\tiny{\textcolor{red}{(-0.6)}} & 41.1~\tiny{\textcolor{red}{(-0.5)}} \\
        {No global pool} & 4.48 & 683 &  82.0~\tiny{\textcolor{red}{(-0.3)}} & 45.8~\tiny{\textcolor{red}{(-0.4)}} & 41.2~\tiny{\textcolor{red}{(-0.4)}} \\
        {Top-only} & 4.49 & 698 & 81.9~\tiny{\textcolor{red}{(-0.4)}} & 45.7~\tiny{\textcolor{red}{(-0.5)}} & 41.2~\tiny{\textcolor{red}{(-0.4)}} \\
        {No gating} & 4.48 & 707 & 81.9~\tiny{\textcolor{red}{(-0.4)}} & 45.6~\tiny{\textcolor{red}{(-0.6)}} & 41.1~\tiny{\textcolor{red}{(-0.5)}} \\
        \bottomrule
    \end{tabular}
    }
    \captionsetup{font=footnotesize} 
    \caption{Component analysis for focal modulation. Four separate changes are made to the original FocalNet. Throughput is reported on image classification. All variants have almost the same size (28.6M) as the default model.}
    \label{tab:ablation_study}
    \vspace{-4mm}
\end{minipage}
\quad
\begin{minipage}{0.45\linewidth}
\setlength{\tabcolsep}{1.8pt}
\resizebox{0.99\linewidth}{!}{
    \footnotesize
    \centering
    \begin{tabular}{l|ccccc}
    \toprule
       Levels (Kernels) & \makecell{Receptive \\ Field} & \#Param. & FLOPs & Throughput & {Top-1} \\
       \toprule
        2 (3-5)     & 7  & 28.4 & 4.41  & 743 & 82.1  \\
        3 (3-5-7)   & 13 & 28.6 & 4.49  & 696 & 82.3  \\
        \midrule
        0 (n/a)     & 0  & 28.3 & 4.35  & 883 & 75.7   \\
        1 (3)       & 3  & 28.3 & 4.37  & 815 & 82.0   \\
        4 (3-5-7-9) & 21 & 29.0 & 4.59  & 592  & 82.2   \\
        \midrule
        1 (13)      & 13 & 28.8 & 4.59  & 661 & 81.9   \\
        \bottomrule
    \end{tabular}
    }
    \captionsetup{font=footnotesize}    
    \caption{Model performance with number of focal levels $L$. ``Receptive Field'' refers to effective receptive field at the top level regardless of the global average pooling.}
    \label{tab:aggregation_level}    
    \vspace{-4mm}
\end{minipage}
\vspace{-3mm}
\end{table}


\textbf{Component Analysis}. Here we ablate FocalNet to study the relative contribution of each component. The result is reported in Table~\ref{tab:ablation_study}, where we investigate the impact of the following model architecture changes on model performance:
\begin{itemize}[leftmargin=*]
    \item \textbf{Replacing Multiplication with Addition}: we change the element-wise multiplication to addition in Eq.~\eqref{Eq:FocalModulation}, which converts the modulator into a bias term. This leads to 0.7\% accuracy drop, which indicates that element-wise multiplication is a more powerful way of modulation than addition.
    \item \textbf{No Global Aggregation}: we remove the top global average pooling in focal modulation. It hurts the performance by $0.3\%$. Even though the hierarchical aggregation already covers a relatively large receptive field, global information ($\Zmat^{L+1}$) is still useful for capturing global context.
    \item \textbf{Top-only Aggregation}: Instead of aggregating the feature maps from all focal levels, we only use the top level map. In this case, the features at lower levels that are more ``local'' and ``fine-grained'' are completely discarded. This change leads to 0.4\% performance drop, which verifies our hypothesis that features at different levels and spatial scopes compensate each other.
    \item \textbf{None-gating Aggregation}: We remove the gating mechanism when aggregating the multiple levels of feature maps. This causes 0.4\% drop. As we discussed earlier, the dependencies between visual token (query) and its surroundings differ based on the query content. The proposed gating mechanism helps the model to {\it adaptively} learn where and how much to gather.
\end{itemize}

In Table~\ref{tab:aggregation_level}, we study the effect of varying the focal level (\emph{i.e.} the number of depth-wise convolution layers $L$). In our experiments reported above, the results show that large receptive field in general achieves better performance (LRF \textit{v.s.} SRF). Here, we investigate by further altering $L$. In additional to setting $L=2$ and $3$, we also try $L=0$, $L=1$, and $L=4$. Accordingly, increasing $L$ brings slight improvement and finally reaches a plateau. Surprisingly, a single level with kernel size 3 can already obtain a decent performance. When we increase the single-level kernel size from 3 to 13, there is a slight 0.1\% drop, and a 0.4\% gap to the one with three levels but same size of receptive field (second row). This indicates that simply increasing the receptive field does not necessarily improve the performance, and a hierarchical aggregation for both fine- and coarse-grained contexts is crucial. 

\subsection{Comparisons with ViTs and ConvNeXts}

\begin{figure}
    \centering
    \includegraphics[width=1.0\linewidth]{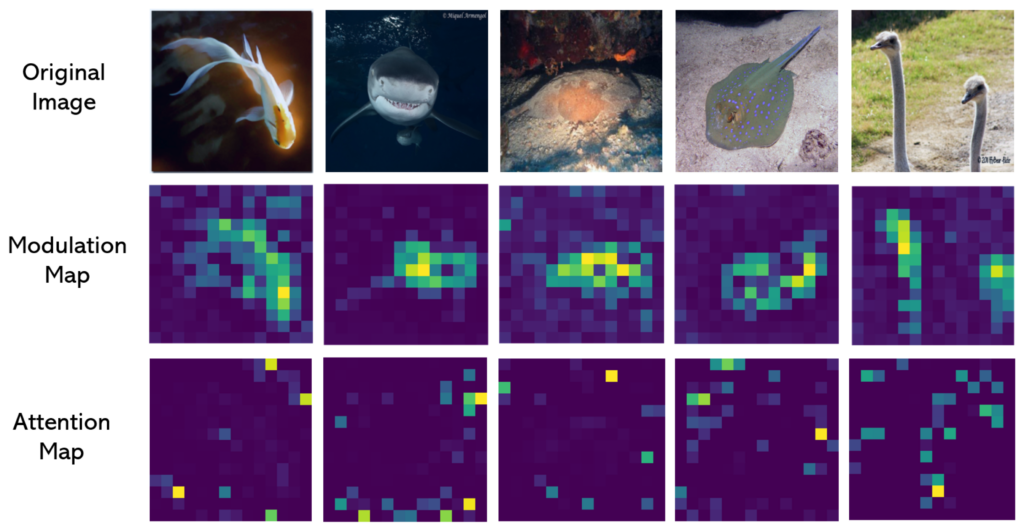}
    \caption{Given an image, we compare the modulation map in our FocalNet-B/16 and the attention map for ViT-B/16. Clearlt, our FocalNet exhibits stronger interpretability than ViT in that it can precisely locate the foreground objects.}
    \label{fig:modulation_attention_maps}
\end{figure}

\begin{table}[t]
\resizebox{0.98\linewidth}{!}{
\footnotesize
    \begin{tabular}{l|cccc|cc|cc|ccc|ccc}
        \toprule
        & \multicolumn{6}{|c}{\makecell{Image Classification}} & \multicolumn{5}{|c}{\makecell{Object Detection}} & \multicolumn{3}{|c}{\makecell{Segmentation}} \\
        \midrule
        \multirow{2}{*}{Model}    &  \multicolumn{4}{|c}{\makecell{Multi-scale}} & \multicolumn{2}{c}{\makecell{Monolithic}} & \multicolumn{2}{|c}{\makecell{Mask R-CNN}} & \multicolumn{3}{c|}{\makecell{C. Mask R-CNN}} & \multicolumn{3}{c}{\makecell{UperNet}} \\
         & Tiny & Small & Base & Large & Small & Base & \multicolumn{2}{c|}{Tiny 3$\times$} &  \multicolumn{3}{c|}{Tiny 3$\times$} & Tiny & Small & Base \\        
        \midrule
        Metric & \multicolumn{4}{c|}{Top-1 Acc.} & \multicolumn{2}{c|}{Top-1 Acc.} & AP$^b$ & AP$^m$ &  AP$^b$ &  AP$^b_{50}$ &  AP$^b_{75}$& \multicolumn{3}{c}{mIoU} \\        
        \midrule
        ConvNeXt~\cite{liu2022convnet}  & 82.1 & 83.1 & 83.8 & \textbf{86.6} & 79.7 & 82.0 & 46.2 & 41.7 & 50.4 & 69.1 & 54.8 & 46.7 & 49.6 & 49.9 \\
        \rowcolor{Gray}   
        FocalNet~(Ours) & \textbf{82.3} & \textbf{83.5} & \textbf{83.9} & 86.5 & \textbf{80.9} & \textbf{82.4} & \textbf{47.6} & \textbf{42.6} & \textbf{51.5} & \textbf{70.1} & \textbf{55.8} & \textbf{47.2} & \textbf{50.1} & \textbf{51.1} \\
        \bottomrule
    \end{tabular}
    }
    \captionsetup{font=footnotesize}
    \caption{Comparison with ConvNeXts with compiled results on a range of computer vision tasks. The numbers of ConvNeXt are reported in~\cite{liu2022convnet}.}
    \label{tab:compare_to_convnext}
    \vspace{-7mm}
\end{table}

ViTs~\cite{dosovitskiy2020image} is undoubtedly the seminal work that applies Self-Attention for visual modeling. Here, we trace back to the origin and study whether our Focal Modulation can fit the monolithic architectures like ViTs. We replace all SA modules in ViTs with focal modulation to construct monolithic FocalNet-T/S/B. We use patch size 16 and three focal levels with kernel sizes 3,5 and 7, so that the effective receptive field is close to the global SA in ViT. 
\begin{wraptable}{r}{5.8cm}
\vspace{-10pt}
\footnotesize
\resizebox{1.0\linewidth}{!}{
\setlength\tabcolsep{2.0pt}
    \begin{tabular}{l|ccccl}
    \toprule
       Model  & Dim & \#Param. & FLOPs & Th.~(imgs/s) & Top-1  \\
       \midrule
       ViT-T/16 & 192 & 5.7 & 1.3 & 2834 & 72.2 \\
    \rowcolor{Gray} 
       FocalNet-T/16 & 192 & 5.9 & 1.1 & 2334 & \textbf{74.1}~{\textcolor{green!50!black}{(+1.9)}} \\
       ViT-S/16 & 384 & 22.1 & 4.6 & 1060 & 79.9\\
     \rowcolor{Gray} 
       FocalNet-S/16 & 384 & 22.4 & 4.3 & 920 & \textbf{80.9}~{\textcolor{green!50!black}{(+1.0)}} \\
       ViT-B/16   & 768 & 86.6 & 17.6 & 330 & 81.8 \\
    \rowcolor{Gray} 
       FocalNet-B/16 & 768 & 87.2 & 16.9 & 300 & \textbf{82.4}~{\textcolor{green!50!black}{(+0.6)}} \\     
       \bottomrule
    \end{tabular}
    }
    \vspace{-5pt}
    \captionsetup{font=footnotesize}    
    \caption{Comparisons between FocalNet and ViT both with monolithic architectures.}
    \vspace{-8pt}
    \label{tab:isotropic}
\end{wraptable}
As shown in Table~\ref{tab:isotropic}, FocalNets consistently outperform ViTs, with similar FLOPs and speed. Besides the quantitative comparisons, we also show the modulation maps and attention maps for FocalNet-B/16 and ViT-B/16 in Fig.~\ref{fig:modulation_attention_maps}. FocalNets clearly demonstrate the stronger interpretability than ViTs.

In Sec.~\ref{Sec:RelatedWork}, we briefly discuss several concurrent works to ours. Among them, ConvNeXts~\cite{liu2022convnet} achieved new SoTA on some challenging vision tasks. Here, we quantitatively compare FocalNets with ConvNeXts by summarizing the results on a series of vision tasks in Table~\ref{tab:compare_to_convnext}. FocalNets outperform ConvNeXts in most cases across the board. Our FocalNets use depth-wise convolution as in ConvNeXt for contextualization but also use modulation to inject the contexts to each individual tokens, which makes a significant difference.

\section{Conclusion}
\vspace{-3pt}
In this paper, we have proposed \emph{Focal Modulation}, a new and generic mechanism that enables input-dependent token interactions for visual modeling. It consists of a hierarchical contextualization to gather for each query token its contexts from short- to long-range, a gated aggregation to adaptively aggregate context features based on the query content into modulator, followed by a simple yet effective modulation. With \emph{Focal Modulation}, we built a series of simple attention-free Focal Modulation Networks (FocalNets) for various vision tasks. Extensive experiments show that FocalNets significantly outperform the SoTA SA counterparts (\eg Swin and Focal Transformer) with similar time-/memory-cost on the tasks of image classification, object detection and semantic segmentation. Notably, our FocalNets achieved new SoTA performance on COCO object detection with much less parameters and pretraining data than the prior works. These encouraging results render Focal Modulation a favorable and even better choice to SA for effective \emph{and} efficient visual modeling.

\textbf{Future works}. The main goal of this work is to develop a more effective way for visual token interaction. Though it seems straightforward, a more comprehensive study is needed to verify whether our Focal Modulation can be applied to other domains such as NLP tasks. Moreover, when coping with multi-modal tasks, SA can be feasibly transformed to cross-attention by alternating the queries and keys. The proposed Focal Modulation requires a gather of contexts for individual queries. How to perform the so-called cross-modulation needs more exploration for multi-modal learning.

\paragraph{Acknowledgement.} We would like to thank Lei Zhang, Hao Zhang, Feng Li and Shilong Liu from IDEA team for helpful discussions and detailed instructions of using DINO for object detection. We would like to thank Aishwarya Kamath for sharing the Object365v2 dataset. We would like to thank Lingchen Meng for helping converting contrastive denoising into regular denoising in DINO.

{\small
\bibliographystyle{plain}
\bibliography{neurips.bib}
}

\newpage
\appendix
\section{More Implementation Details}

\subsection{Model Configuration}

As we discussed in our main submission, we observed in our experiments that different configurations (\textit{e.g.}, depths, dimensions, \textit{etc}) lead to different performance. For a fair comparison, we use the same stage layouts and hidden dimensions as Swin~\cite{liu2021swin,yang2021focal}, but replace the SA modules with Focal Modulation modules. We thus construct a series of Focal Modulation Network (FocalNet) variants as shown in Table~\ref{tab:focalnet_variants}.

\begin{table}[h]
    \centering
\resizebox{0.99\linewidth}{!}{    
    \begin{tabular}{l|cc|ccc}
      \toprule
      Name & Depth & {Dimension ($d$)} & \makecell{Levels \\ ($L$)} & \makecell{Kernel Size \\ ($k^1$)} & \makecell{Effective Receptive Field \\ ($r^L$)} \\
      \midrule
         {FocalNet-T~(SRF/LRF)}  & [2,2,6,2] &  [96,192,384,768] &  &  &  \\
         FocalNet-S~(SRF/LRF) & [2,2,18,2] & [96,192,384,768] & [2,2,2,2] & [3,3,3,3] & [7,7,7,7] \\
         FocalNet-B~(SRF/LRF) & [2,2,18,2] & [128,256,512,1024] & [3,3,3,3] & [3,3,3,3] & [13,13,13,13] \\
         FocalNet-L~(SRF/LRF) & [2,2,18,2] & [192,384,768,1536] & & & \\
         \midrule
         FocalNet-H & [2,2,18,2] & [352,704,1408,2816] & [4,4,4,4] & [3,3,3,3] & [21, 21, 21, 21]\\
         \bottomrule
    \end{tabular}
    }
    \captionsetup{font=footnotesize}  
    \caption{Model configurations at four stages for FocalNet. The depth layouts and hidden dimension ($d$) are the same to Swin~\cite{liu2021swin} and Focal Transformers~\cite{yang2021focal}. SRF and LRF means small and large receptive field, respectively. The only difference is the number of focal levels ($L$) and starting kernel size ($k^{\ell=1}$). The last column lists the effective receptive field at top focal level at each stage ($r^L$).}
    \label{tab:focalnet_variants}
    \vspace{-3mm}
\end{table}

\subsection{Training settings for ImageNet-1K} 
We follow Swin~\cite{liu2021swin} to use the same set of data augmentations including Random Augmentation~\cite{cubuk2020randaugment}, Mixup~\cite{zhang2017mixup}, CutMix~\cite{yun2019cutmix} and Random Erasing~\cite{zhong2020random}. For model regularization, we use Label Smoothing~\cite{szegedy2016rethinking} and DropPath~\cite{huang2016deep}. For all models, the initial learning rate is set to $10^{-3}$ after 20 warm-up epochs beginning with $10^{-6}$. For optimization, we use AdamW~\cite{loshchilov2017decoupled} and a cosine learning rate scheduler~\cite{loshchilov2016sgdr}. The weight decay and the gradient clipping norm is set to $0.05$ and $5.0$, respectively. We set the stochastic depth drop rates to $0.2$, $0.3$ and $0.5$ for our tiny, small and base models, respectively. During training, images are randomly cropped to $224 \times 224$, and a center crop is used during evaluation. Throughput/Speed is measured on one V100 GPU with batch size 128, following~\cite{liu2021swin}. A detailed summary is shown in Table~\ref{tab:config4imagenet1k}.

\begin{table}[h]
    \centering
    \footnotesize
    \begin{tabular}{l | c c}
    \toprule
        Setting & FocalNet-T/S/B (Hierarchical) &
        FocalNet-T/S/B (Monolithic) \\
        \midrule
        batch size          &  1024 & 1024   \\
        base learning rate  &  1e-3 & 1e-3   \\
        learning rate scheduler & cosine & cosine \\
        min learning rate   & 1e-5   & 1e-5 \\
        training epochs     &  300  & 300    \\
        warm-up epochs      &  20   & 20     \\
        warm-up schedule    & linear & linear \\
        warm-up learning rate & 1e-6  & 1e-6 \\
        optimizer          & adamw   & adamw \\
        \midrule
        color jitter factor & 0.4   & 0.4 \\
        auto-aug          & rand-m9-mstd0.5-inc1 & rand-m9-mstd0.5-inc1 \\
        random-erasing prob. & 0.25 & 0.25 \\
        random-erasing mode  & pixel & pixel \\
        mixup  $\alpha$         & 0.8   & 0.8 \\
        cutmix $\alpha$         & 0.8   & 0.8 \\
        mixup prob.             & 1.0   & 1.0 \\
        mixup switch prob.      & 0.5   & 0.5 \\
        \midrule
        stochastic drop path rate & 0.2/0.3/0.5 & 0.2/0.2/0.3 \\
        label smoothing        & 0.1  & 0.1 \\
        gradient clip          & 5.0  & 5.0 \\
        weight decay           & 0.05 & 0.05 \\
        
    \bottomrule
    \end{tabular}
    \caption{Experimental settings for training on ImageNet-1K with FocalNets.}
    \label{tab:config4imagenet1k}
\end{table}

\subsection{Training settings for ImageNet-22K}
We train FocalNet-B and FocalNet-L for 90 epochs with a batch size of 4096 and input resolution $224 \times 224$. The initial learning rate is set to $10^{-3}$ after a warmup of 5 epochs. We set the the stochastic depth drop rates to $0.2$ for both networks. For stability, we use LayerScale~\cite{touvron2021going} with initial value $10^{-4}$ for all layers. The other settings follow those for ImageNet-1K. After the pretraining, we finetune the models on ImageNet-1K for 30 epochs with initial learning rate of $3\times 10^{-5}$, cosine learning rate scheduler and AdamW optimizer. The stochastic depth drop rate is set to $0.3$ and both CutMix and Mixup are muted during the finetuning.

\subsection{Training settings for Object365}

We exactly follow the settings in DINO~\cite{zhang2022dino} to pretrain our object detection models on Object365~\cite{shao2019objects365}. In total, Object365 contains around 1.7M images for training and 80k for validation. We merge 75k validation images to the training data and use the remained for evaluation during the pretraining. We pre-train FocalNets+DINO for 26 epochs with learning rate $1e^{-4}$, and drop the learning rate by 10 times after 24 epochs. A standard image resolution $800 \times 1333$ is used. After the pretraining, we finetune the model on COCO with max size $1200\times 2000$. We lower the learning rate to $6e^{-5}$ and train the model for 12 epochs in total.

\begin{table}[t]
    \centering
    \footnotesize
    \begin{tabular}{l | c c}
    \toprule
        Setting & FocalNet-B/L (Pretraining) &
        FocalNet-B/L (Finetuning) \\
        \midrule
        resolution          &  224$\times$224 & 224$\times$224 and 384$\times$384 \\
        batch size          &  4096 & 1024   \\
        base learning rate  &  1e-3 & 3e-5   \\
        learning rate scheduler & cosine & cosine \\
        min learning rate   & 1e-5   & 5e-6 \\
        training epochs     &  90  & 30    \\
        warm-up epochs      &  5   & 0     \\
        warm-up schedule    & linear & linear \\
        warm-up learning rate & 1e-6  & 1e-6 \\
        optimizer          & adamw   & adamw \\
        \midrule
        color jitter factor & 0.4   & 0.4 \\
        auto-aug          & rand-m9-mstd0.5-inc1 & rand-m9-mstd0.5-inc1 \\
        random-erasing prob. & 0.25 & 0.25 \\
        random-erasing mode  & pixel & pixel \\
        mixup  $\alpha$         & 0.8   & n/a \\
        cutmix $\alpha$         & 0.8   & n/a \\
        mixup prob.             & 1.0   & n/a \\
        mixup switch prob.      & 0.5   & n/a \\
        initial layer scale             & 1e-4  & pretrained \\
        \midrule
        stochastic drop path rate & 0.2/0.2 & 0.3 \\
        label smoothing        & 0.1  & 0.1 \\
        gradient clip          & 5.0  & 5.0 \\
        weight decay           & 0.05 & 1e-8 \\
        
    \bottomrule
    \end{tabular}
    \caption{Experimental settings for pretraining on ImageNet-22K with FocalNet-B/L and finetuning on ImageNet-1K.}
    \label{tab:config4imagenet22k}
\end{table}

\section{Downstream Tasks}
\subsection{Object Detection}
\subsubsection{Effect of kernel size} 
\begin{figure}[t]
\begin{minipage}{0.48\linewidth}
	\centering
	\includegraphics[width=1.0\linewidth]{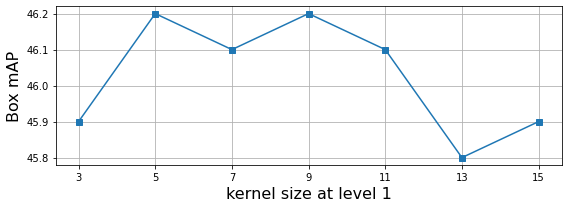}
	\includegraphics[width=1.0\linewidth]{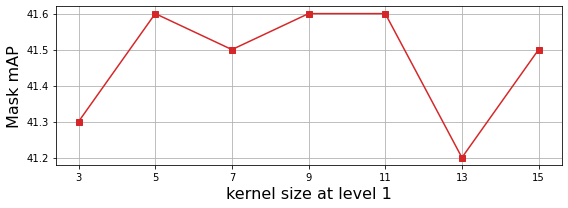}	
	\vspace{-5mm}
    \captionsetup{font=scriptsize}    
    \caption{Box and mask mAP for Mask R-CNN 1$\times$ training. We use FocalNet-T (LRF) as the baseline model and vary its kernel size at first level $k^{\ell=1} \in \{3,5,7,9,11,13,15\}$.}
    \vspace{-3mm}
    \label{fig:boxmask_map_with_ks}	
\end{minipage}
\quad
\begin{minipage}{0.48\linewidth}
	\centering
	\includegraphics[width=1.0\linewidth]{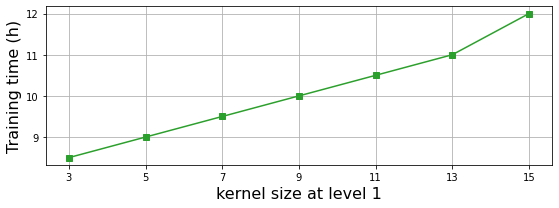}
	\includegraphics[width=1.0\linewidth]{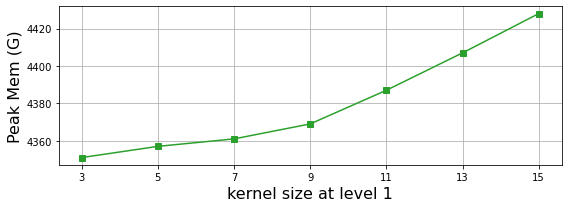}	
	\vspace{-5mm}
    \captionsetup{font=scriptsize}    
    \caption{Training time (wall-clock) and peak memory for Mask R-CNN 1$\times$. We train Focalnet-T (LRF) with different kernel sizes on 16 V100 GPUs with batch size 16.}
    \vspace{-3mm}
    \label{fig:time_mem_with_ks}	
\end{minipage}
\end{figure}

We study how the various kernel sizes affect the object detection performance when finetuning FocalNet-T (LRF) with $k^{\ell=1}=3$ pretrained on ImageNet-1K. In Fig.~\ref{fig:boxmask_map_with_ks}, we vary the kernel size at first level $k^{\ell=1}$ from 3 to 15 for object detection finetuning. 
We have two interesting observations: $(i)$ though the pretrained model used $k^{\ell=1}=3$, it can be finetuned with different kernel sizes to adapt high-resolution object detection task; $(ii)$ a moderate kernel size (5,7,9,11) have a slightly better performance than a kernel size which is too small (3) or too big (13,15), probably because small kernel cannot capture the long-range dependency while big kernel misses the detailed local context. In Fig.~\ref{fig:time_mem_with_ks}, we further show the corresponding wall-clock time cost and peak memory when training on 16 V100 GPUs with batch size 16. Accordingly, increasing the kernel size gradually increases the training memory and time cost. For a good performance/cost trade-off, we therefore set $k^{\ell=1}=9$ for all the object detection finetuning experiments in our main submission.

\begin{table}[t]
\centering
\resizebox{0.95\linewidth}{!}{
\setlength{\tabcolsep}{1.9pt}
\footnotesize
\begin{tabular}{lcc|lccccc|lccccc}
\toprule
\multirow{2}{*}{Backbone} & \#Params & FLOPs & \multicolumn{6}{c}{Mask R-CNN 1x} & \multicolumn{6}{c}{Mask R-CNN 3x}\\
\cmidrule{4-15}
 & (M) & (G) & $AP^b$ & $AP^b_{50}$ & $AP^b_{75}$ & $AP^m$ & $AP^m_{50}$ & $AP^m_{75}$ & $AP^b$ & $AP^b_{50}$ & $AP^b_{75}$ & $AP^m$ & $AP^m_{50}$ & $AP^m_{75}$\\
\midrule
FocalNet-T~(SRF) & 48.6 & 267  & {45.9} & {68.3} & {50.1} & {41.3} & {65.0} & {44.3} & {47.6} & {69.5} & {52.0} & {42.6} & {66.5} & {45.6} \\
FocalNet-T~(LRF) & 48.9 & 268 & {46.1} & {68.2} & {50.6} & {41.5} & {65.1} & {44.5} & {48.0} & {69.7} & {53.0} & {42.9} & {66.5} & {46.1} \\
\rowcolor{Gray}
FocalNet-T~(SRF)$\dagger$ & 45.8 & 261  & {46.8} & {69.1} & {51.2} & {41.9} & {65.6} & {44.6} & {48.5} & {70.0} & {53.2} & {43.3} & {67.0} & {46.3} \\
\midrule
FocalNet-S~(SRF) & 70.8 & 356 & {48.0} & {69.9} & {52.7} & {42.7} & {66.7} & {45.7} &  {48.9} & {70.1} & {53.7} & {43.6} & 67.1 & {47.1} \\
FocalNet-S~(LRF) & 72.3 & 365 & {48.3} & {70.5} & {53.1} & {43.1} & {67.4} & {46.2} & {49.3} & {70.7} & {54.2} & {43.8} & {67.9} & {47.4} \\
\rowcolor{Gray}
FocalNet-S~(SRF)$\dagger$ & 59.5 & 312 & {48.1} & {70.5} & {52.8} & {43.1} & {67.2} & {46.2} & {49.2} & {70.6} & {53.9} & {43.8} & {67.6} & {47.2}  \\

\midrule
FocalNet-B~(SRF) & 109.4 & 496 &  {48.8} & {70.7} & {53.5} & {43.3} & {67.5} & {46.5} & {49.6} & {70.6} & {54.1} & {44.1} & {68.0} & {47.2} \\
FocalNet-B~(LRF) & 111.4 & 507  & {49.0} & {70.9} & {53.9} & {43.5} & {67.9} & {46.7} & {49.8} & {70.9} & {54.6} & {44.1} & {68.2} & {47.2} \\
\rowcolor{Gray}
FocalNet-B~(SRF)$\dagger$ & 107.1 & 481  & {49.6} & {71.2} & {54.6} & {44.0} & {68.2} & {47.6} & {50.2} & {71.0} & {55.0} & {44.3} & {68.1} & {47.9} \\
\bottomrule
\end{tabular}}
\captionsetup{font=footnotesize}    
\caption{Gray rows are additional results using deeper but thinner FocalNets in Table~\ref{tab:deeper_network} as the backbone.}
\label{tab:maskrcnn_deepernet}
\end{table}

\subsubsection{Results with deeper and thinner FocalNets} 
In our main submission, we compared with previous SoTA methods Swin and Focal Transformer in a restricted way by using the same network depth layout. Meanwhile, we also showed that different depth layouts lead to different image classification performance. Here, we investigate how the layout affects the object detection performance. We use the deeper but thinner FocalNets in Table 4 of our main submission as the backbones. Specifically, we change the depth layout of our FocalNet-T from 2-2-6-2 to 3-3-16-3, and FocalNet-S/B from 2-2-18-2 to 4-4-28-4. Meanwhile, we reduce the initial hidden dimension from 96, 128 to 64, 96, respectively.
In Table~\ref{tab:maskrcnn_deepernet}, we add the additional gray rows to compare with the results reported in our main submission. In Table~\ref{tab:od_deepernet_lrf}, we further show the 1$\times$ results of deeper and thinner FocalNets with large receptive field. Accordingly, the object detection performance (both box and mask mAP) are boosted over the shallower and wider version of FocalNets with same receptive field. On one hand, this trend suggests a feasible way to improve the performance for our FocalNet, and further demonstrate its effectiveness for both image classification and object detection. \textbf{On the other hand, it suggests that keeping network configuration (depth, hidden dimension, \textit{etc}.) the same is important for a fair comparison with previous works.}

\subsection{Image Segmentation}

In Table~\ref{tab:semantic_seg_deepernet}, we report the results using the deeper and thinner FocalNets as the backbone for semantic segmentation. As we can see, for FocalNet-T, increasing the depth does not bring extra improvement. For larger models, however, a deeper version outperforms the shallow ones, particularly on FocalNet-B. Additionally, we further compare with most recent work MPViT~\cite{lee2021mpvit} which also exploits multi-scale features but in parallel manner. As we can see, our FocalNets achieve better performance than MPViT with comparable cost. Compared with MPViT, the hierarchical and gated contextualization proposed in FocalNets can rapidly cover large receptive field facilitating the high-resolution dense prediction tasks. 

\begin{table}[t]
\begin{minipage}{0.47\linewidth}
\centering
\footnotesize
\resizebox{1.0\linewidth}{!}{
\begin{tabular}{l|cccc}
    \toprule
     Backbone & \#Param. & FLOPs & AP$^b$ & AP$^m$ \\
     \midrule
     Swin-Tiny & 47.8 & 264 & 43.7 & 39.8\\
     FocalAtt-Tiny & 48.8 & 291 & 44.8 & 41.0 \\
     FocalNet-T (SRF) & 48.6 & 267 & {45.9} & {41.3} \\
     FocalNet-T (SRF)$\dagger$ & 45.8 & 261 & {46.8} & {41.9} \\
     FocalNet-T (LRF) & 48.9 & 268 & {46.1} & {41.5} \\
    \rowcolor{Gray}
     FocalNet-T (LRF)$\dagger$ & 46.1 & 262 & {46.7} & {41.9} \\
     \midrule
     Swin-Small & 69.1 & 354 & 46.5 & 42.1 \\
     FocalAtt-Small & 71.2 & 401 & 47.4 & 42.8 \\
     FocalNet-S (SRF) & 70.8 & 356 & {48.0} & 42.7 \\
     FocalNet-S (SRF)$\dagger$ & 59.5 & 312 & {48.1} & {43.1} \\
     FocalNet-S (LRF) & 72.3 & 365 & {48.3} & {43.1} \\
    \rowcolor{Gray}
     FocalNet-S (LRF)$\dagger$ & 60.0 & 315 & {48.6} & {43.3} \\
     \midrule
     Swin-Base & 107.1 & 497 & 46.9 & 42.3 \\
     FocalAtt-Base & 110.0 & 533 & 47.8 & 43.3 \\
     FocalNet-B (SRF) & 109.4 & 496 & {48.8} & {43.3} \\
     FocalNet-B (SRF)$\dagger$ & 107.1 & 481 & {49.6} & {44.0} \\
     FocalNet-B (LRF) & 111.4 & 507 & {49.0} & {43.5} \\
    \rowcolor{Gray}
     FocalNet-B (LRF)$\dagger$ & 107.9 & 485 & {49.9} & {44.2} \\
     \bottomrule
\end{tabular}
}
\captionsetup{font=scriptsize}    
\caption{Additional results of Mask R-CNN 1$\times$ with deeper and thinner FocalNets~(LRF) in gray rows. We use the same pretrained model as FocalNet~(SRF)$\dagger$, but add an extra focal level on top with kernel initialized with all-zeros.}
\vspace{-5mm}
\label{tab:od_deepernet_lrf}
\end{minipage}
\quad
\begin{minipage}{0.52\linewidth}
\centering
\footnotesize
\resizebox{0.98\linewidth}{!}{
  \begin{tabular}{l|ccccc}
    \toprule
    Backbone  & \#Param. & FLOPs & mIoU & +MS \\
    \midrule	 
    Swin-T~\cite{liu2021swin} & 60 & 941 & 44.5 & 45.8  \\
    FocalAtt-T~\cite{yang2021focal} & 62 & 998 & 45.8 & 47.0 \\
    FocalNet-T~(SRF) & 61 & 944 & {{46.5}} & {47.2} \\    
    FocalNet-T~(LRF) & 61 & 949 & {46.8} & {47.8} \\
    \rowcolor{Gray}   
    FocalNet-T~(SRF)$\dagger$ & 55 & 934 & {{47.4}} & {48.5} \\       
    \midrule
    Swin-S~\cite{liu2021swin} & 81 & 1038 & 47.6 & 49.5  \\
    FocalAtt-S~\cite{yang2021focal} &  85 & 1130 & 48.0 & {50.0}\\     
    MPViT-S~\cite{lee2021mpvit} & 52 & 943 & 48.3 & n/a \\    
    FocalNet-S~(SRF) &  83 & 1035 & {49.3} & {50.1} \\     
    FocalNet-S~(LRF) & 84 & 1044 & {{49.1}} & 50.1 \\  
    \rowcolor{Gray}   
    FocalNet-S~(SRF)$\dagger$ &  69 & 986 & {49.4} & {50.3} \\       
    \midrule
    Swin-B~\cite{liu2021swin} &  121 & 1188 & 48.1 & 49.7  \\    
    FocalAtt-B~\cite{yang2021focal} & 126 & 1354 & 49.0 & 50.5 \\ MPViT-B~\cite{lee2021mpvit} & 105 & 1186 & 50.3 & n/a \\
    FocalNet-B~(SRF) & 124 & 1180 & {50.2} & {51.1} \\    
    FocalNet-B~(LRF) &  126 & 1192 & {50.5} & {51.4} \\    
    \rowcolor{Gray}    
    FocalNet-B~(SRF)$\dagger$ & 117 & 1159 & {51.0} & {51.9} \\       
    \bottomrule
  \end{tabular} 
  }
  \captionsetup{font=scriptsize}    
  \caption{Semantic segmentation on ADE20K~\cite{zhou2017scene}. All models are trained with UperNet~\cite{xiao2018unified}. Grays rows are additional results with deeper yet thinner FocalNets (SRF).}
  \label{tab:semantic_seg_deepernet}    
  \vspace{-5mm}
\end{minipage}
\end{table}

\section{Additional Model Interpretation}

Our focal modulation consists of three main components: $(i)$ convolution for contextualization; $(ii)$ gating mechanism for aggregation of multiple granularity and $(iii)$ linear projection for generating modulator. Here we attempt to interpret each of them.

\begin{figure}[t]
\begin{minipage}{0.48\linewidth}
	\centering
	\includegraphics[width=1.05\linewidth]{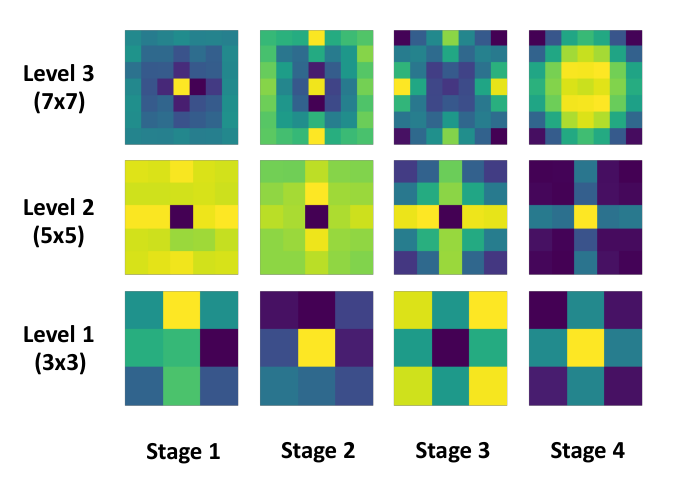}
	\vspace{-3mm}
    \captionsetup{font=scriptsize}    
    \caption{Visualization of learned kernels at three levels and four stages in FocalNet-T~(LRF). For clarity, we only show for the last layer of each stage.}
    \label{fig:focalnet-tiny-lrf-kernels}
\end{minipage}
\quad
\begin{minipage}{0.48\linewidth}
	\centering
	\includegraphics[width=1.05\linewidth]{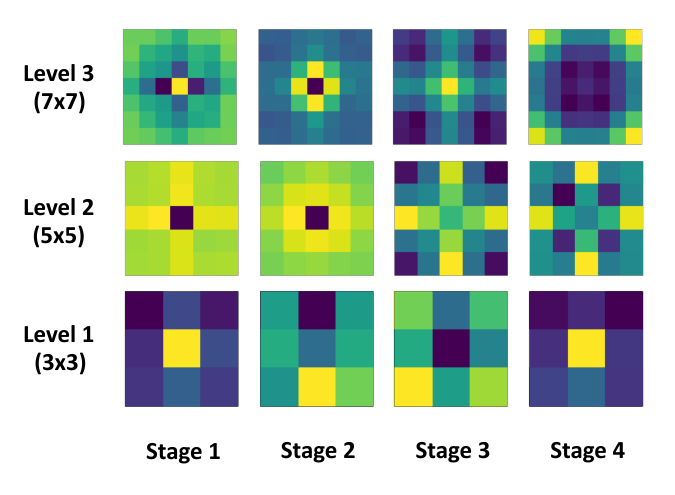}
	\vspace{-3mm}
    \captionsetup{font=scriptsize}    
    \caption{Visualization of learned kernels at three levels and four stages in FocalNet-B~(LRF). For clarity, we only show for the last layer at each stage.}
    \label{fig:focalnet-base-lrf-kernels}
\end{minipage}
\end{figure}

\paragraph{Convolutional kernel patterns at different levels and layers.} In Fig.~\ref{fig:focalnet-tiny-lrf-kernels} and Fig.~\ref{fig:focalnet-base-lrf-kernels}, we show the learned depth-wise convolutional kernels in our FocalNet-T (LRF) and FocalNet-B (LRF). Specifically, we show the averaged 3$\times$3, 5$\times$5 and 7$\times$7 kernels at last layer of each of four stages. We observe some interesting patterns from the visualizations. In the earlier stage, the models usually focus on local regions and thus have more scattered weights at low focal levels (level 1 and 2). Nevertheless, when it comes to later stage, the model requires more global context to make the final prediction, which explains the more scattered weights at the third focal level.

\begin{figure}[t]
    \centering
    \includegraphics[width=1.0\linewidth]{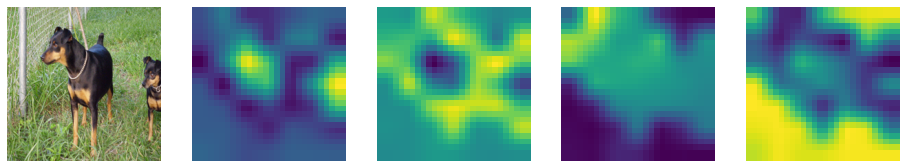}
    \includegraphics[width=1.0\linewidth]{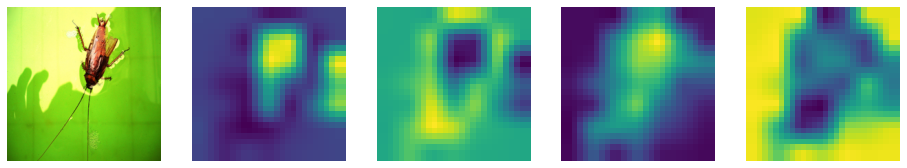}
    \includegraphics[width=1.0\linewidth]{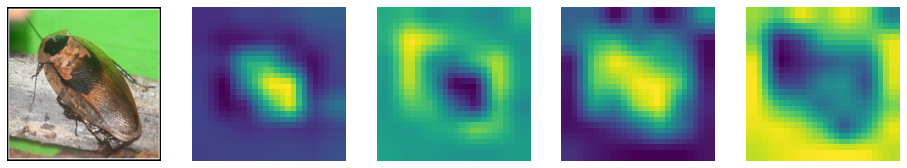}
    \includegraphics[width=1.0\linewidth]{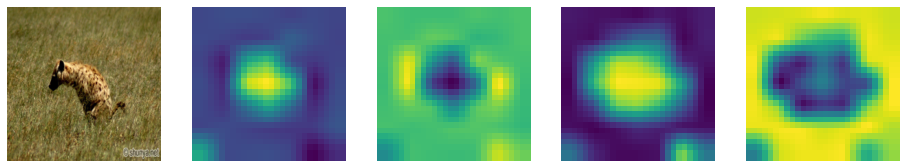}
    \includegraphics[width=1.0\linewidth]{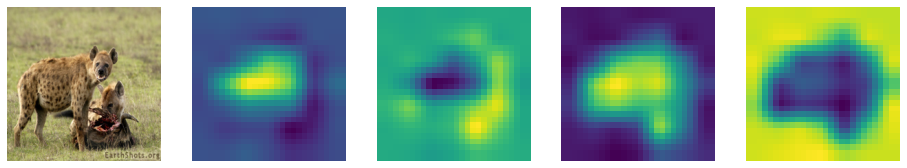}
    \includegraphics[width=1.0\linewidth]{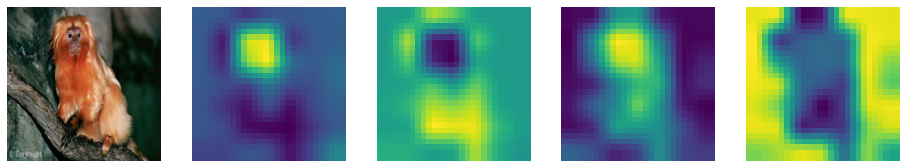}
    \includegraphics[width=1.0\linewidth]{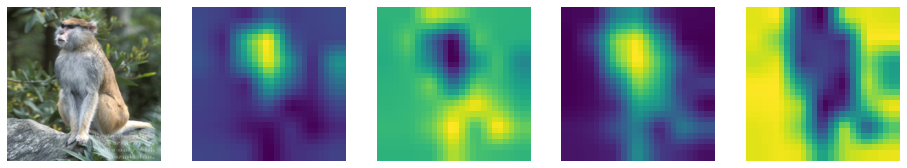}
    \caption{Visualization of gating values $\Gmat$ at last layer of our FocalNet-B~(LRF) pretrained on ImageNet-1K. From left to right, we show input image, and gating weights $\Gmat^{\ell}, \ell = 1,2,3,4$.}
    \label{fig:gating_vis1}
\end{figure}

\begin{figure}[t]
    \centering
    \includegraphics[width=1.0\linewidth]{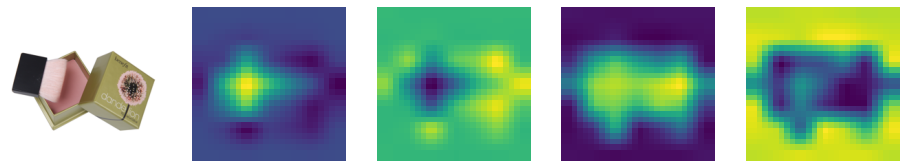}
    \includegraphics[width=1.0\linewidth]{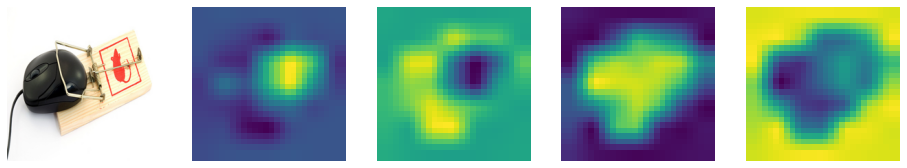}
    \includegraphics[width=1.0\linewidth]{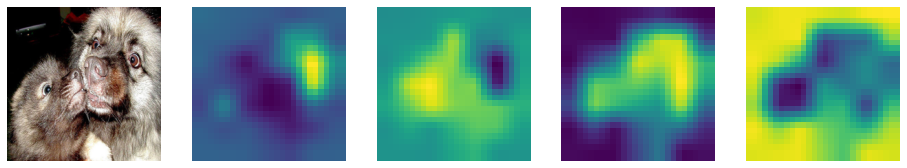}
    \includegraphics[width=1.0\linewidth]{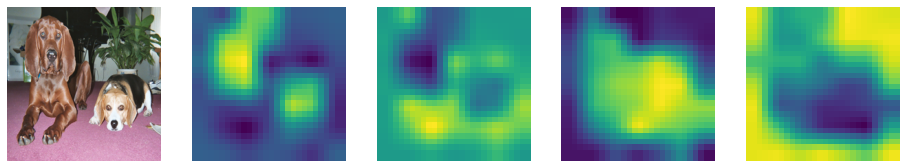}
    \includegraphics[width=1.0\linewidth]{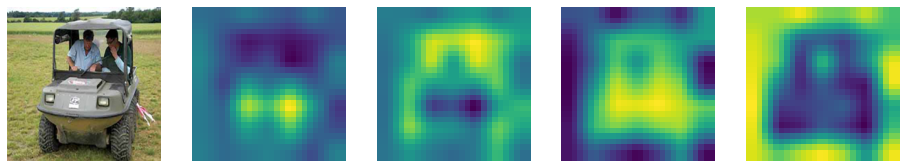}
    \includegraphics[width=1.0\linewidth]{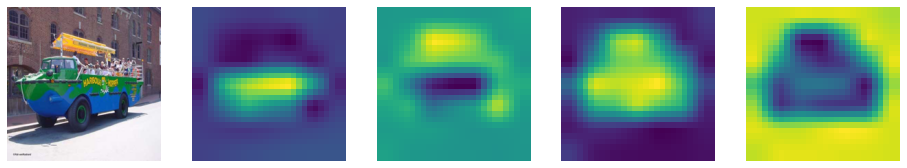}
    \includegraphics[width=1.0\linewidth]{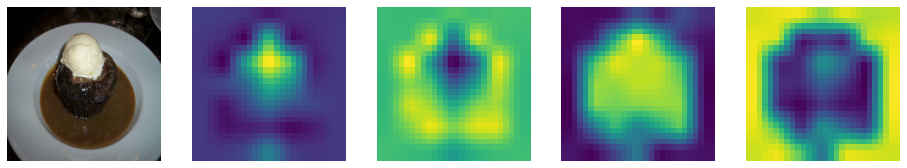}
    \caption{Visualization of gating values $\Gmat$ at last layer of our FocalNet-B~(LRF) pretrained on ImageNet-1K. The order from left to right column is same to Fig.~\ref{fig:gating_vis1}}
    \label{fig:gating_vis2}
\end{figure}

\begin{figure}[t]
    \centering
    \includegraphics[width=1.0\linewidth]{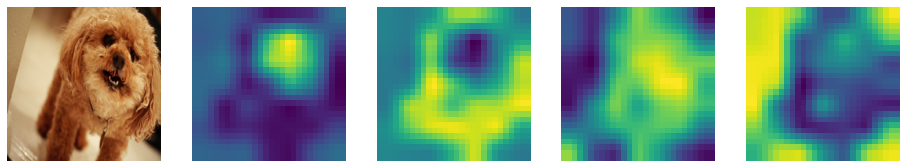}
    \includegraphics[width=1.0\linewidth]{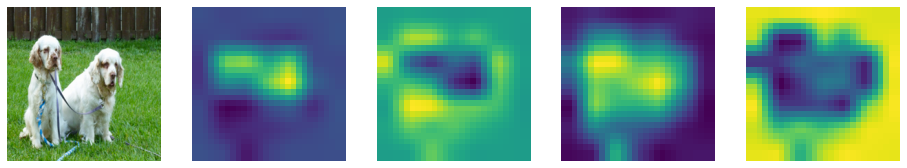}
    \includegraphics[width=1.0\linewidth]{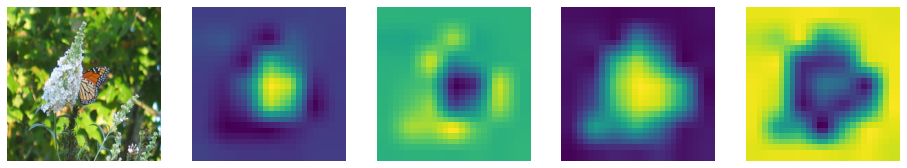}
    \includegraphics[width=1.0\linewidth]{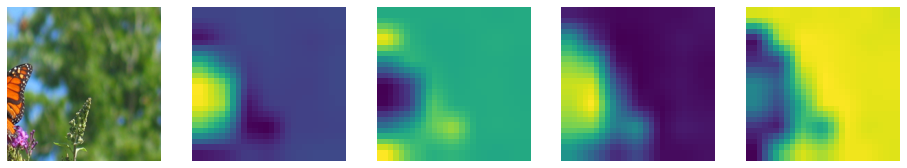}
    \includegraphics[width=1.0\linewidth]{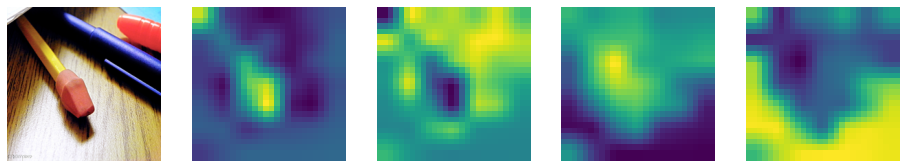}
    \includegraphics[width=1.0\linewidth]{figures/pics/pic23.png}
    \includegraphics[width=1.0\linewidth]{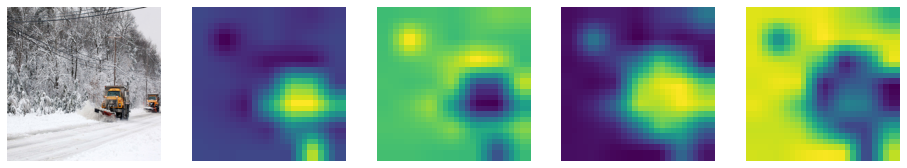}
    \caption{Visualization of gating values ${\Gmat}$ at last layer of our FocalNet-B~(LRF) pretrained on ImageNet-1K. The order from left to right column is same to Fig.~\ref{fig:gating_vis1}}
    \label{fig:gating_vis3}
\end{figure}

\paragraph{Gating function for adaptive contextualization.} Similar to Fig.~\ref{fig:gating_vis_main}, we make more visualizations of the gating values in our FocalNets. On a set of randomly selected ImageNet-1K validation images, we show more gating maps in Fig.~\ref{fig:gating_vis1},~\ref{fig:gating_vis2} and~\ref{fig:gating_vis3}. For the visual tokens at object regions  ($\ell=1$), their gating values are much higher than those outside object regions at first level. When looking more closely, we can see that the predicted gating values mainly lie on the most complicated textures within object regions. At the second level $\ell=2$, the gating values are still higher in object regions but the peak values usually move to the object boundaries instead. At the third level  $\ell=3$, the whole object regions have higher gating values than background regions. Finally at level $\ell=4$, we find there is a clear distinction between foreground and background regions when aggregating the global contexts. The foreground regions usually show less interest in the global context and the other way around for the background regions. Even for those images containing multiple foreground objects, our model still shows coherent patterns. Comparing the gating values for first three levels and the last global context, we can find our model does gather more information from local regions when modulating foreground visual tokens and more global context for background tokens. 

\begin{figure}[t]
    \centering
    \includegraphics[width=0.49\linewidth]{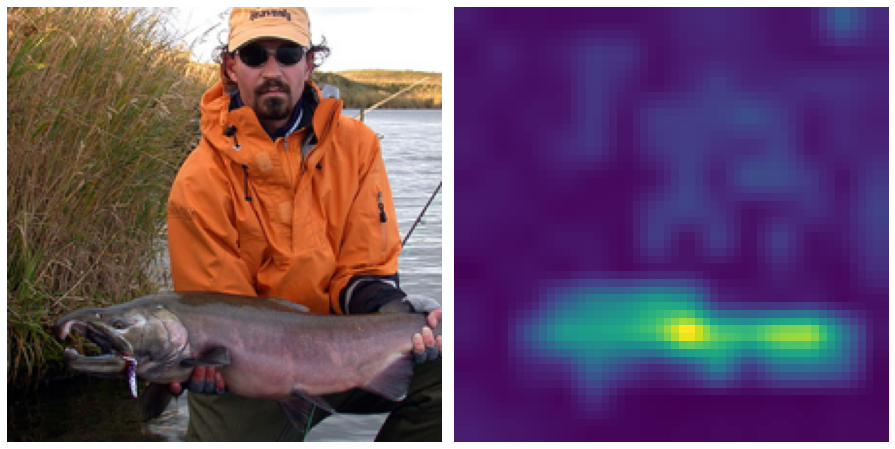}
    \includegraphics[width=0.49\linewidth]{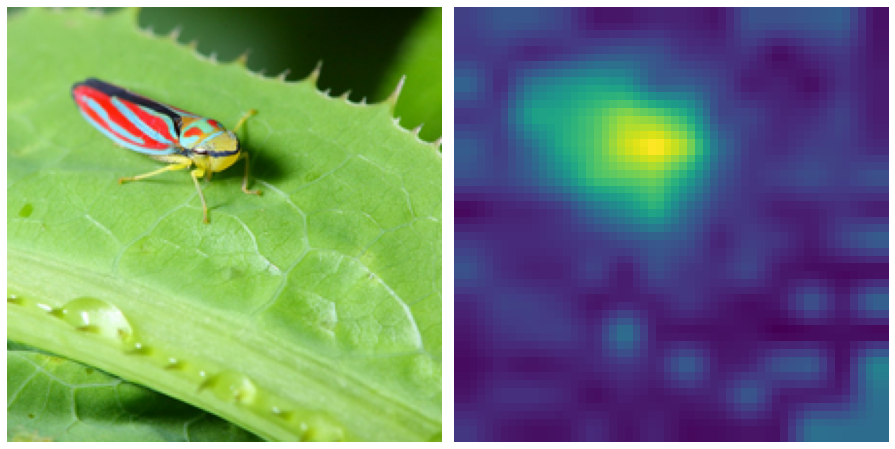}

    \includegraphics[width=0.49\linewidth]{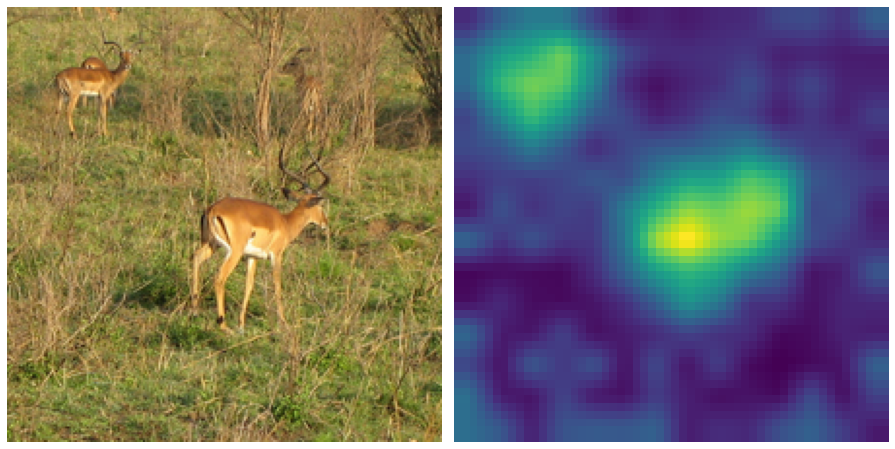}
    \includegraphics[width=0.49\linewidth]{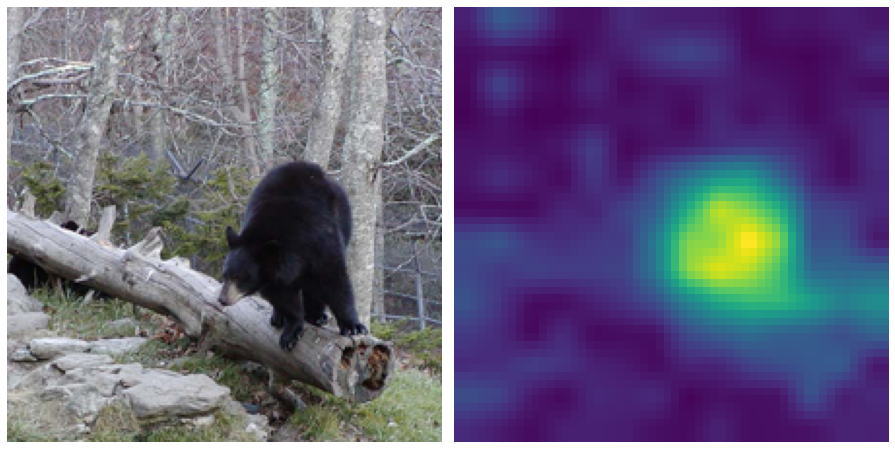}

    \includegraphics[width=0.49\linewidth]{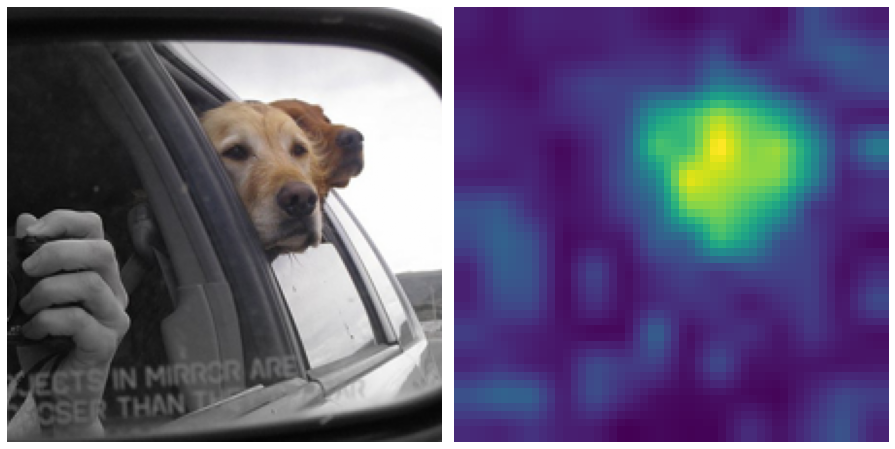}
    \includegraphics[width=0.49\linewidth]{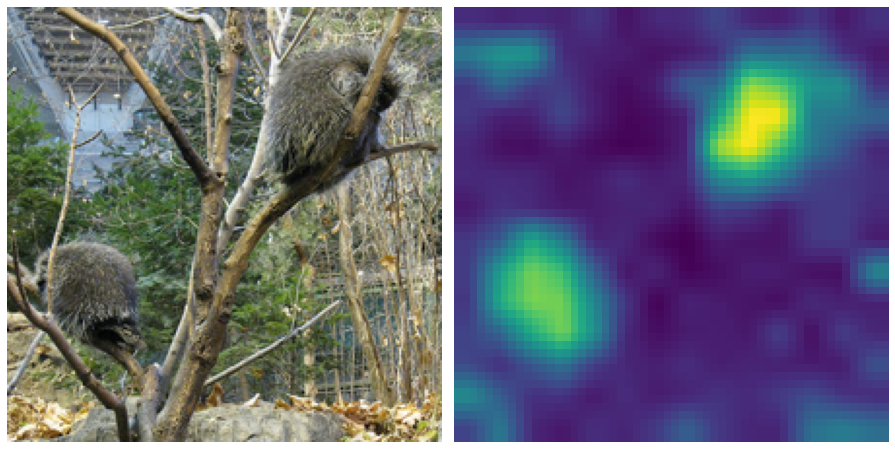}
    
    \includegraphics[width=0.49\linewidth]{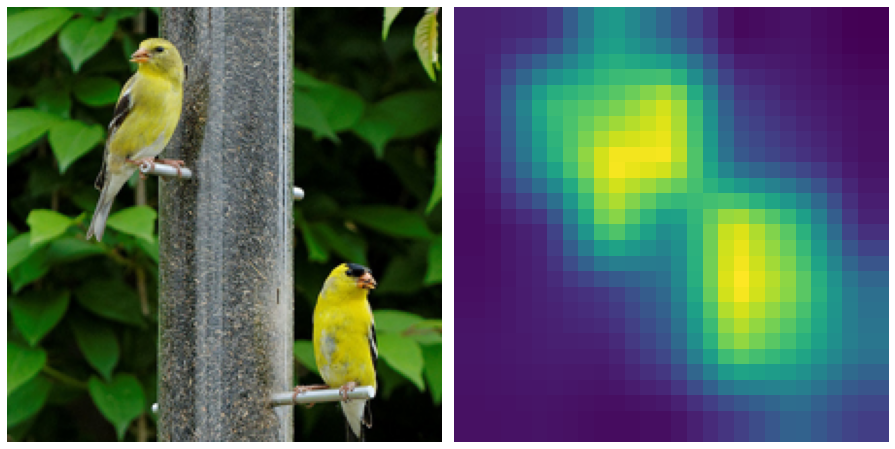}
    \includegraphics[width=0.49\linewidth]{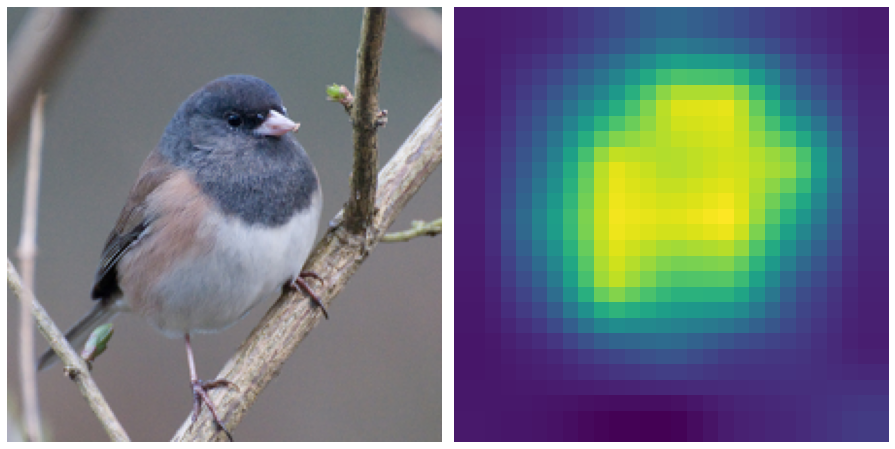}

    \includegraphics[width=0.49\linewidth]{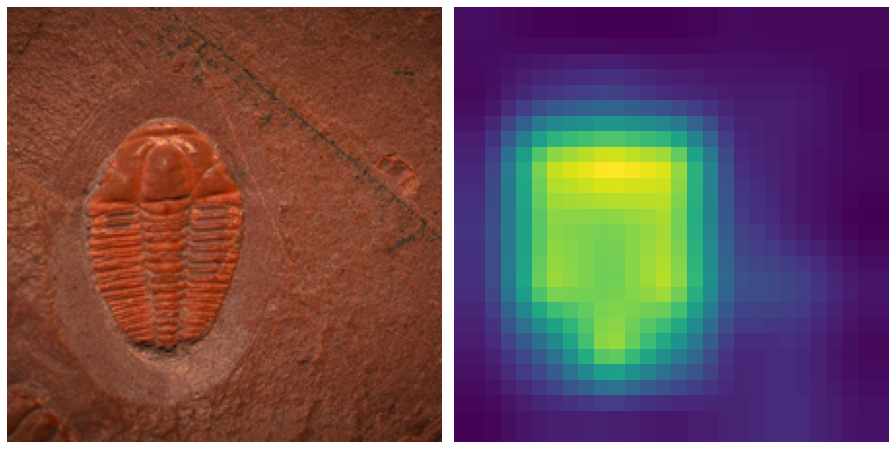}
    \includegraphics[width=0.49\linewidth]{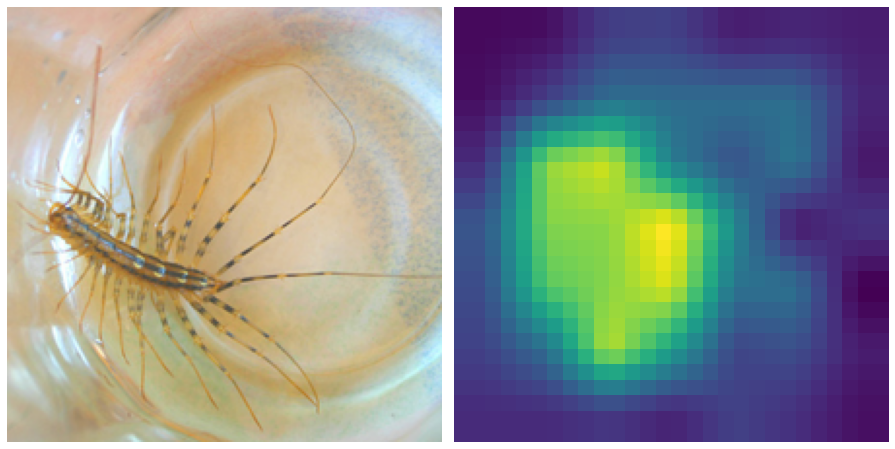}

    \includegraphics[width=0.49\linewidth]{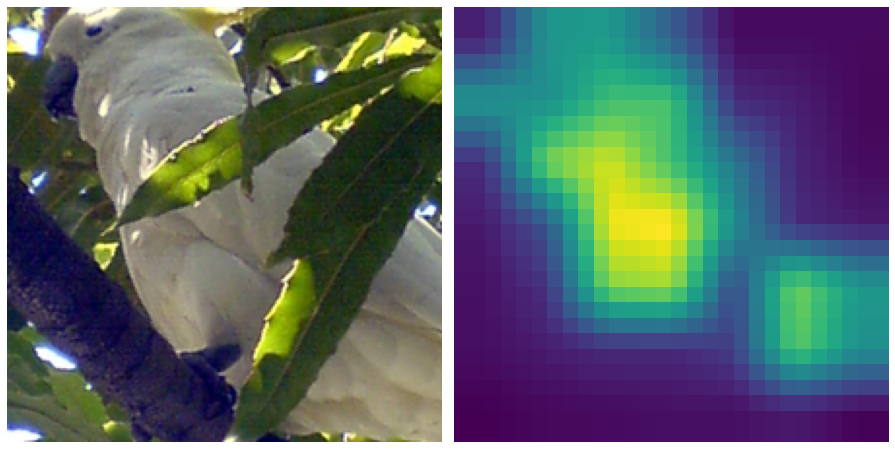}
    \includegraphics[width=0.49\linewidth]{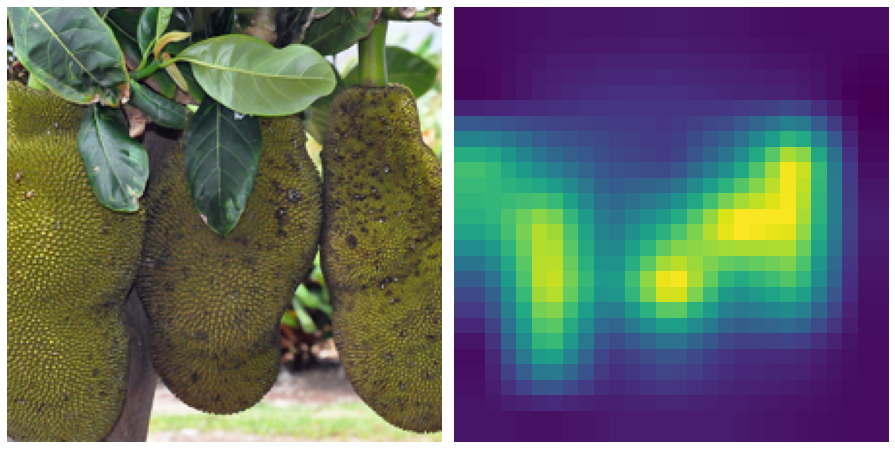}    

    \caption{Visualization of modulator at the top layer of FocalNets. We show the magnitude ($L$-2 norm) of the modulator vector pixel-wise for each input image. The heatmaps clearly show the ``attention'' of our FocalNets for free. The top three rows are with the monolithical FocalNet-B/16 and the bottom three rows are for the hierarchical FocalNet-Base.}
    \label{fig:modulator_supp}
\end{figure}

\paragraph{Modulator is a new way of attention.} As we showed earlier, the modulator $\Mmat$ computed at the last layer of our FocalNet shows an emerge of object localization capacity even though no class guidance is provided, unlike CAM~\cite{zhou2016learning} or Grad-CAM~\cite{selvaraju2017grad}. We show additional visualizations in Fig.~\ref{fig:modulator_supp}. They clearly show that our FocalNets is good at localizing the most discriminative regions from the image and thus spending more effort to modulate these regions to produce the final predictions. We strongly believe this property emerged in our proposed Focal Modulation mechanism opens up a new door for the community on how to interpret the models beyond the gradient-based and class-guided methods. We will leave studies on the correlations between the modulators and the correctness of final predictions, and the robustness of our FocalNets as future works. We refer the readers to try our online modulator visualization demo at \url{https://huggingface.co/spaces/jw2yang/focalnet-modulators}.

\section{Social Impact}


This work is mainly focused on architecture design for computer vision tasks. We have trained the models on various datasets and tasks. One concern is that it might be biased to the training data. When it is trained on large-scale webly-crawled image data, the negative impact might be amplified due to the potential offensive or biased contents in the data. To avoid this, we need to have a careful sanity check on the training data and the model's predictions before training the model and deploying it to the realistic applications.




\end{document}